\documentclass{article}

    \PassOptionsToPackage{numbers, compress}{natbib}

 \usepackage[main, final]{neurips_2024}

\usepackage{graphicx}
\usepackage{makecell}
\usepackage[utf8]{inputenc} 
\usepackage[T1]{fontenc}    
\usepackage{hyperref}       
\usepackage{url}            
\usepackage{booktabs}       
\usepackage{amsfonts}       
\usepackage{nicefrac}       
\usepackage{microtype}      
\usepackage{enumitem}
\usepackage{multicol}
\usepackage{hhline}
\usepackage{kotex}
\usepackage[capitalize]{cleveref}
\usepackage{multirow}  
\usepackage[table]{xcolor}

\usepackage[T1]{fontenc}
\usepackage{courier}   
\usepackage{upquote}   
\usepackage{subcaption}
\usepackage{arydshln}
\title{\includegraphics[height=1.0em]{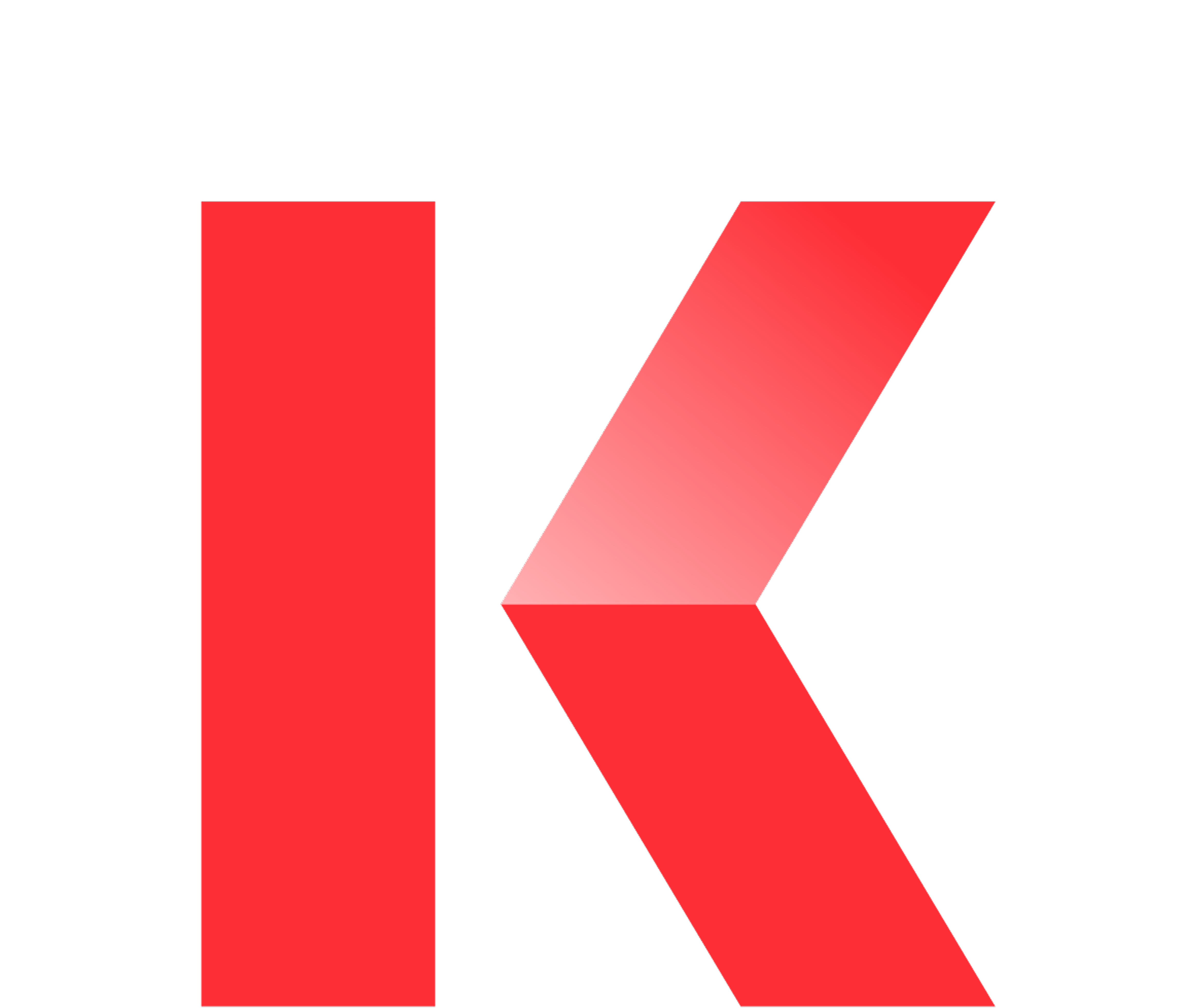}
Mi:dm K 2.5 Pro}

\author{Tech. Innovation Group, KT \\
        midm-llm@kt.com}

\begin{document}
\maketitle
\begin{abstract}
The evolving LLM landscape requires capabilities beyond simple text generation, prioritizing multi-step reasoning, long-context understanding, and agentic workflows. This shift challenges existing models in enterprise environments, especially in Korean-language and domain-specific scenarios where scaling is insufficient. We introduce Mi:dm K 2.5 Pro, a 32B parameter flagship LLM designed to address enterprise-grade complexity through reasoning-focused optimization. 

Our methodology builds a robust data foundation via a quality-centric curation pipeline utilizing abstract syntax tree (AST) analysis for code, gap-filling synthesis for mathematics, and an LLM-based quality evaluator. Pre-training scales the model via layer-predictor-based Depth Upscaling (DuS) and a progressive strategy supporting a 128K token context window. Post-training introduces a specialized multi-stage pipeline—including Reasoning SFT, model merging, and asynchronous reinforcement learning (RL)—to develop complex problem-solving skills. "Fusion Training" then rebalances these capabilities with conversational fluency, consistent response styling, and reliable tool-use. 

The evaluations show that Mi:dm K 2.5 Pro achieves competitive performance against leading global and domestic models. In addition, it sets state-of-the-art results on Korean-specific benchmarks, showcasing deep linguistic and cultural understanding. Finally, Responsible AI evaluations validate safety against attacks, ensuring a secure profile for deployment with a balance of harmlessness and responsiveness.
\end{abstract}
\section{Introduction}
\label{section1:introduction}

\begin{figure}[h!]
  \centering
  \includegraphics[width=0.95\linewidth]{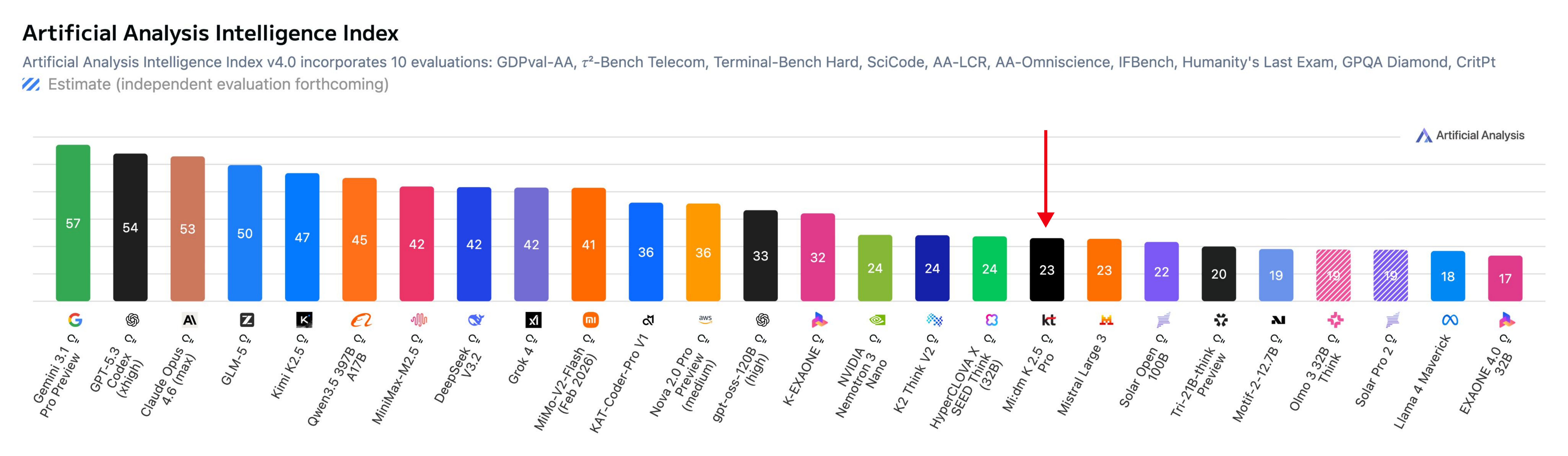}
   \caption{Artificial Analysis Intelligence Index (AAII) v4.0 results}
   \label{fig:2-6-1}
\end{figure}

KT introduces Mi:dm K 2.5 Pro, the flagship model in the Mi:dm K large language model (LLM) lineup. Building on the technical foundations validated by the existing Mi:dm 2.0 Base and Mini models, this report provides a comprehensive account of the Pro model’s training pipeline and performance, enabled by increased parameter count and reasoning-focused optimization.

Recent developments in the LLM landscape increasingly require capabilities beyond simple question answering (QA) and text generation, with advanced multi-step reasoning, long-context understanding, external tool use, and agentic workflows emerging as key evaluation criteria~\cite{aaii_2025}. This shift reflects the evolution of LLMs from response generators into more sophisticated systems capable of complex decision-making and problem solving.

However, existing small to mid scale models exhibit limitations in complex reasoning, long-context analysis, and reliable instruction following required in enterprise environments, due to their constrained parameter counts and context windows~\cite{wei2022emergentabilitieslargelanguage}. This challenge is especially pronounced for Korean-language, real-world industrial data and domain-specific scenarios, where simple model scaling alone is often insufficient to achieve the required levels of reliability and performance.

With this motivation, Mi:dm K 2.5 Pro is developed with explicit design objectives aimed at addressing enterprise-grade complexity, rather than through naive model scaling alone. The model is configured with 32B parameters—approximately 3× larger than Mi:dm 2.0 Base~\cite{shin2026mi}—to increase knowledge density and support deeper reasoning. In addition, we extend the context length from 32K to 128K tokens, enabling robust handling of enterprise requirements such as long-document analysis, multi-document understanding, and complex information extraction.

Moving beyond simple scaling, Mi:dm K 2.5 Pro can dynamically adapt its inference behavior to balance reasoning-intensive tasks with latency-sensitive general requests~\cite{Guo_2025}. This enables robust reasoning for complex computation and problem-solving, while supporting efficient response generation for general-purpose informational queries and natural conversation.

From a data perspective, KT leveraged extensive public and private sector partnerships to curate proprietary domain datasets that are difficult to obtain through open-web document collection alone. Constructed under a sophisticated data governance framework and an advanced preprocessing and quality-filtering pipeline, the resulting training corpus strengthens Korean-language understanding and improves coverage of Korea-centric context, thereby differentiating the model.

We subdivide the post-training pipeline and perform phase-specific reward modeling and hyperparameter tuning aligned with the objective of each stage. By incorporating model merging~\cite{wortsman2022model}, we improve training stability and achieve balanced performance across complex reasoning, coding, instruction following, and agentic task execution.

To ensure a consistent and reliable user experience, we define the Mi:dm K response style guide and systematically strengthen persona consistency, tone and register control, and response-structure stability during SFT. As a result, Mi:dm K 2.5 Pro maintains robust instruction-following performance under multiple constraints and on highly complex tasks.

Furthermore, we have expanded Mi:dm K from a Korean-English bilingual model to a multilingual model supporting four languages, including Japanese and Chinese. Using high-quality language-specific data during training, we aim to generate responses that reflect each language's context and cultural characteristics, rather than relying on simple translation.

With the development of Mi:dm K 2.5 Pro, KT has established a complete Mi:dm K lineup spanning Mini, Base, and Pro. This enables flexible model selection based on available compute and target use cases. Building on this lineup, KT is committed to serving as a core engine for accelerating customers’ AI transformation (AX) by delivering industry-specific AI solutions.
\section{Data Foundations}
\label{section2:data}
The data strategy of Mi:dm K 2.5 Pro builds on the Korean-language knowledge base established in Mi:dm 2.0~\cite{shin2026mi}, shifting the emphasis from scaling data volume to quality-centric curation and effective utilization. While Mi:dm 2.0 primarily targeted broad, general-purpose Korean competence, Mi:dm K 2.5 Pro seeks to improve domain-specific reasoning and real-world usability through rigorous data selection and systematic, structure-aware data restructuring.

To this end, we adopt a data curation pipeline that integrates an LLM-based data quality evaluator with abstract syntax tree (AST) analysis, instead of relying on heuristic filters. For domains where data are sparse or exhibit structural biases—such as mathematics and code—we augment the training distribution with synthetic data. In addition, we define a Mi:dm K response-style specification and apply rewrite-based post-processing to ensure consistent output formatting and robust adherence to instruction. In this section, we describe the data acquisition and curation procedures under this strategy.

\subsection{High Quality Data Acquisition}             
\label{subsection2-1:data}
Mi:dm K 2.5 Pro builds upon the training data used for Mi:dm 2.0~\cite{shin2026mi}, while expanding domain coverage and rebalancing the data mixture. This expansion has three objectives: (i) augmenting Korean-language training data, (ii) selectively incorporating multilingual data, and (iii) broadening training coverage for STEM, code, and agentic domains.

We source training data through three channels: licensed proprietary datasets, public datasets that permit commercial use, and in-house synthetic data. Accordingly, as we scale the corpus, rigorous curation—refinement and selection—becomes critical to prevent quality degradation.

\paragraph{Korean Data Acquisition Strategy.}            
Mi:dm K 2.5 Pro maintains a Korean-centric training strategy and constructs its training corpus primarily from public resources, including AI-HUB\footnote{\url{https://www.aihub.or.kr}} and the National Institute of Korean Language (NIKL) corpora\footnote{\url{https://www.korean.go.kr}}. In addition, through sustained collaborations with an industry–academia, including the \textit{K-Data Alliance} —we secure large volumes of Korea-focused data that reflect the institutional context, cultural norms, and knowledge landscape of Korean society.

The collected corpus encompasses materials rooted in Korean socio-economic and institutional contexts—covering areas such as economy and finance, public administration, as well as cultural and social domains—together with Korean-language specialized knowledge, including STEM education content, expertise distilled from professional publications, and code- and agent-oriented scenarios. Instead of being siloed for narrow applications, these data are integrated into the general pretraining mixture, strengthening both general linguistic competence and domain-specific reasoning capabilities.

We apply the same refinement pipeline used for the existing training corpus to the newly collected Korean-language source data, and utilize a subset of it for synthetic data generation. The synthesized data help enrich Korean-language knowledge coverage in specialized domains where real-world resources are relatively scarce.

\paragraph{Multilingual Support and Language Transfer.} 
To support foundational multilingual capability while preserving a Korean-centric learning focus, we intentionally limit the multilingual data to 3–10\% of the Korean corpus. This controlled inclusion is designed to encourage beneficial cross-lingual transfer without diluting the primary emphasis on Korean-language training.

We construct the multilingual corpus using Chinese and Japanese. For Chinese, we sample web documents from the top-scoring, ultra-high-quality band of the OpenCSG Chinese Corpus~\cite{yu2025opencsg} and Cosmopedia-format synthetic documents~\cite{benallal2024cosmopedia} generated from the same corpus. For Japanese, we use documents from the Japanese subset of FineWeb2-HQ~\cite{messmer2025multilingdatacomp} that pass our internal toxicity filters.
In addition, we curate web-based knowledge resources in Chinese and Japanese, including them after applying the same refinement and selection criteria.
Beyond raw multilingual corpora, we further construct translation datasets derived from the source multilingual data. Translations are generated and stratified by task type (e.g., question answering, information extraction, summarization) and by translation direction (e.g., en $\leftrightarrow$ zh, ja $\leftrightarrow$ ko). During training, translated data are mixed with their corresponding source-language data to promote cross-lingual alignment while maintaining balanced exposure.

Across all multilingual and translated data, we extend our existing refinement pipeline with language-aware deduplication and heuristic filtering tailored to language-specific characteristics. We also apply toxicity filtering to minimize culturally or socially sensitive content, taking into account the model’s intended use in Korean-centric training and deployment contexts. Overall, multilingual data serve a complementary role and are carefully controlled to support—rather than override—the Korean-focused training objective.

\paragraph{Specialized Domain (STEM, Code, and Agentic).}
To expand coverage in STEM, code, and agent-oriented (tool-use) tasks, we use publicly available specialist datasets such as peS2o~\cite{peS2o} and OpenStax\footnote{\url{https://openstax.org/}}. In addition, we acquire problem-bank resources and code datasets through commercial licensing and partnerships.

\subsection{Refinement Pipeline for Code}
\label{subsection2-2:data-code}
\begin{figure}[h!]
  \centering
  \includegraphics[width=0.8\linewidth]{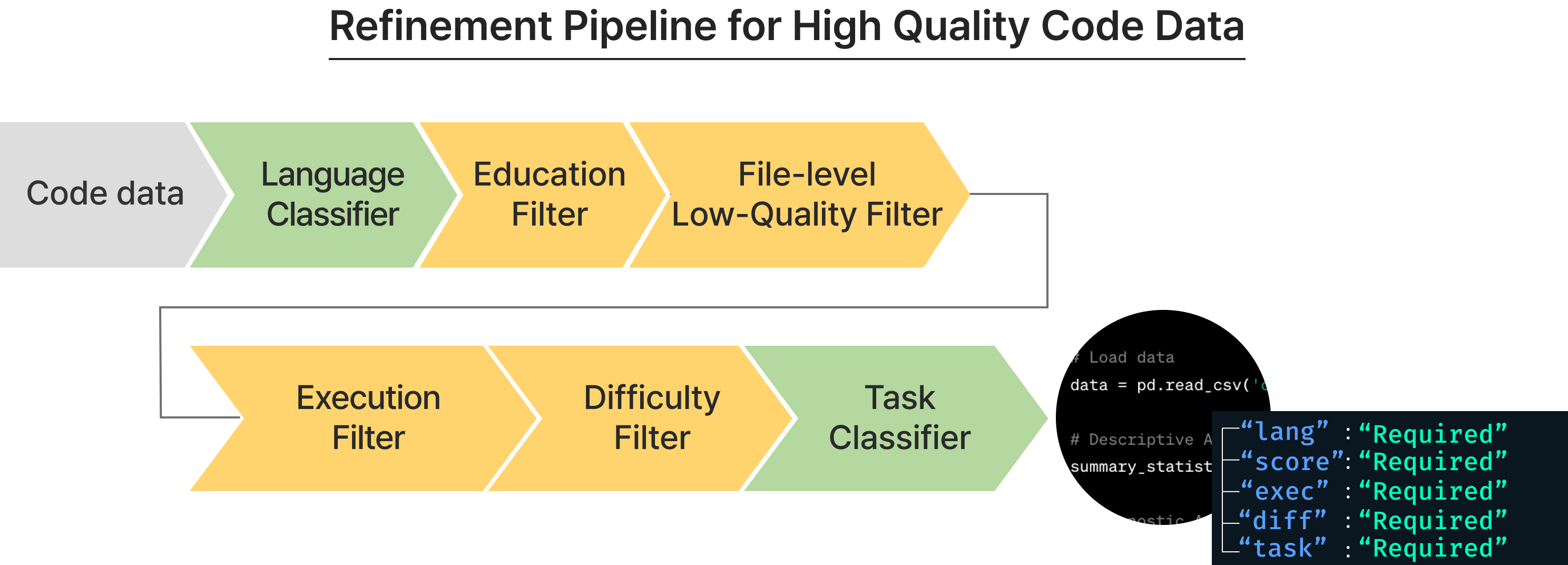}
   \caption{Refinement pipeline for code}          
   \label{fig:code-pipeline}
\end{figure}

To effectively leverage code reasoning data for training, it is necessary to go beyond coarse tagging (e.g., difficulty, language, or pass/fail correctness). We should systematically filter out training-incompatible samples while constructing the remaining corpus to broadly reflect real-world usage across programming languages, difficulty tiers, and programming task types.

Accordingly, we apply a sequential filtering and annotation pipeline to the input code corpus consisting of:
(i) programming-language identification $\rightarrow$
(ii) educational suitability and quality criteria $\rightarrow$
(iii) removal of source- and file-level spurious artifacts $\rightarrow$
(iv) executability-based filtering $\rightarrow$
(v) difficulty labeling $\rightarrow$
(vi) task-type labeling.
Each sample is annotated with structured metadata, including {\small \texttt{language}, \texttt{quality\_score}, \texttt{is\_executable}, \texttt{difficulty}}, and {\small\texttt{task}}. These metadata serves as core axes for sampling policies and mixture scheduling during multi-stage LLM training.

\paragraph{Programming Language Classifier.}           
Code data exhibits substantial structural variation—for example, natural language and code may be interleaved within a single sample, and code fences may explicitly specify the target language—making misclassification common when relying on a single criterion. To address this, we adopt a two-step programming-language identification strategy.

First, if the code fence specifies a language in the form {\small\verb|```{language}|}, we use it as the primary signal for language assignment. Otherwise, we infer the language using a dedicated classifier. We define a predefined set of widely used languages based on real-world prevalence, and map all remaining instances to an {\small\texttt{other}} category.

In our comparison between lexing-based tools (e.g., Pygments) and model-based approaches for programming-language classification, we find that lexing-based methods tend to be unreliable when discriminating between syntactically similar languages. By contrast, the model-based classifier provides more consistent and higher accuracy on major languages. We therefore adopt the model-based approach as the primary signal for language identification in our pipeline.

\paragraph{Education Score Filter.}
From a code-learning perspective, samples that perform no substantive logical operations, are not self-contained or executable, or consist solely of high-level instructions provide limited supervision for code reasoning. Accordingly, we extend our existing quality assessment criteria into a 5-point training-suitability scale (1–5) and remove low-scoring samples during the refinement stage.

Samples with scores of 1–3 typically lack executability or explicit reasoning, featuring incomplete code, syntax errors, or snippets too brief to demonstrate problem-solving logic. In contrast, samples in the 4–5 range contain executable, self-contained code and make the problem-solving logic explicit, including reasoning steps such as computation and conditional logic.

In our pipeline, samples with scores of 1–3 are treated as training-incompatible and filtered out, while samples with scores of 4–5 are retained to support the development of code reasoning ability. The goal of this refinement step is not to deliver short-term gains in answer accuracy, but to raise the quality floor of the training corpus, which in turn improves the reliability of the difficulty and task labels assigned in subsequent stages.

\paragraph{File-Level Low Quality Filter.}          
Recurring noise from the same generation pipeline or upstream source is inefficient to address via per-sample inspection alone. We therefore use a rule-based, file-level detector to identify systematically recurring noise patterns and drop the corresponding files/sources in bulk.

Concretely, we exclude files during refinement when the natural-language proportion is so high that code-type classification becomes unstable, the training-suitability scores of many samples within the same file are consistently low, or multiple unrelated functions are interleaved such that the training unit of an individual sample is unclear.

\paragraph{Execution Filter.}
For code reasoning data, not only answer correctness but also structural validity and executability of the code are crucial. To this end, the final pipeline includes a execution filter that parses source code into an abstract syntax tree (AST) and assesses executability at the syntactic and structural levels. Compared with simple string-based heuristics, this approach more reliably captures structural properties of code.

The execution filter distinguishes genuinely executable code from samples that merely resemble code blocks but are in fact corrupted fragments, mixed natural-language/code text, or formats unsuitable for execution testing. This filter serves as a critical curation step in constructing high-quality code data and helps improve the reliability of the downstream task classifier that identifies code execution tasks.

\paragraph{Difficulty Filter.}
In model training, difficulty informs the overall data strategy, particularly the design of sampling and mixture composition. In the final stage, we categorize code difficulty into three levels: Easy, Medium, and Hard. Difficulty is defined based on the types of algorithms and data structures involved, as summarized in~\cref{tab:difficulty_by_algorithm}. For example, problems centered on arrays or hash tables are labeled Easy; those involving recursion, heaps, dynamic programming, or graph algorithms are labeled Medium; and problems requiring more advanced techniques—such as segment trees, Fenwick trees (BIT), advanced dynamic programming, or optimization methods—are labeled Hard.

This process does not rely on a single feature (e.g., code length); instead, it bases difficulty on the sophistication of the conceptual tools required for problem solving. This criterion is particularly important for reasoning data: as rationales become longer, they may include redundant or non-essential explanation, making length-based difficulty estimation unreliable.

\begin{table}[t]
\centering
\small
\renewcommand{\arraystretch}{1.15}
\setlength{\tabcolsep}{8pt}
\resizebox{\textwidth}{!}{
\begin{tabular}{
>{\raggedright\arraybackslash}p{2.5cm}
>{\raggedright\arraybackslash}p{12.5cm}
}
\toprule\midrule\rowcolor{gray!10}

\textbf{Difficulty} & \textbf{Algorithm Category} \\       
\midrule
Easy &
Array, String, Hash Table, Math, Simulation, $\cdots$ \\ \hdashline
Medium &
Binary Search, Sliding Window, Greedy, Heap, Backtracking, Topological Sort, Union-Find, Tree/Graph, DP, $\cdots$ \\\hdashline
Hard &
Suffix Array, Aho--Corasick, Min-Cost Max-Flow, Heavy-Light Decomposition, Li Chao Tree, Convex Hull Trick, Matrix Exponentiation, Digit/Tree DP, $\cdots$ \\
\bottomrule
\end{tabular}
}
\caption{Difficulty classification based on algorithm categories}            
\label{tab:difficulty_by_algorithm}
\end{table}

\paragraph{Task Classifier.}

Reasoning data in the code domain spans multiple tasks beyond Code Generation, including Self-Repair, Test Output Prediction, and Code Execution. The task distribution can materially affect training dynamics~\cite{wang2025aethercode, jain2024livecodebench, zheng2024top}. To reflect the diverse coding capabilities required in real-world use cases, we clearly define each task and adopt a criteria-driven classification scheme. Moreover, to handle boundary cases—e.g., samples that already include problem-solving code being misclassified as Code Generation—we apply a combination of prompt-level rules and post-processing rules to maintain label consistency.

\paragraph{Code Data Distribution Analysis.}            
Finally, each sample is a code instance annotated with metadata including {\small\texttt{language}}, {\small\texttt{quality\_score}}, an {\small\texttt{is\_executable}} flag (AST-based executability), {\small\texttt{difficulty}}, and {\small\texttt{task}}. Rather than serving merely as descriptive annotations, these metadata are actively used to guide filtering decisions and to control sampling ratios and distribution balancing across training stages. For example, if task-distribution analysis reveals an excessive proportion of Self-Repair tasks, we adjust sampling weights in subsequent stages to match a target distribution. Conversely, for tasks with relatively low proportion, we design additional data collection or targeted data generation strategies to restore balance.

\begin{figure}[h!]
\centering
\includegraphics[width=0.9\linewidth]{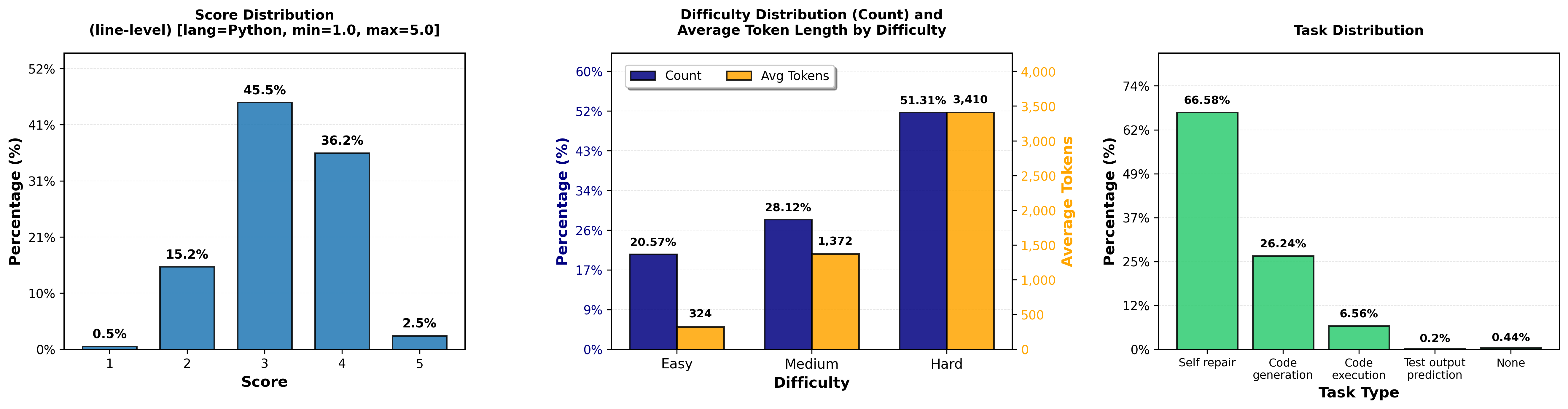}
\caption{Distribution report of {\small\texttt{score, difficulty}}, and {\small\texttt{task}} after refinement (distribution bias diagnosis based on top buckets).}
\label{fig:code-score-dist}
\end{figure}

Building on these classification results, we remove contaminated samples and construct high-quality data buckets, thereby reducing error propagation in the staged pipeline (language $\rightarrow$ quality $\rightarrow$ is\_executable $\rightarrow$ difficulty $\rightarrow$ task) and preventing upstream mistakes from cascading into downstream labels.
In addition, coupling an AST-based executability filter with task classification provides verifiability at the data layer, a key requirement for code-domain training. As a result, the {\small\texttt{language/quality\_score/executable/difficulty/task}} meta tags enable stage- and objective-aware data selection and mixture control.

\subsection{Structural Refinement and Synthesis for Math}
\label{subsection2-3:data-math}

Unlike general knowledge or code data, the difficulty of mathematical problems is determined not only by the concepts involved but also by the number of reasoning steps required to apply those concepts. In other words, even problems based on simple concepts can become highly challenging when they demand multi-step reasoning~\cite{webb1997criteria}. As a result, the quality of mathematical data cannot be adequately characterized through surface-level curation alone.

To address this, we design a structured refinement pipeline that analyzes data distributions across domain, conceptual level, and reasoning depth, leveraging these signals to guide the direction of data synthesis. Specifically, we structure the mathematical data along these three primary axes to ensure a balanced and high-quality training corpus.

The domain axis represents the mathematical subject area of each problem. Following the Mathematics Subject Classification (MSC) 2020~\cite{dunne2020mathematics}, we map 63 fine-grained categories into seven high-level domains: Algebra, Geometry \& Topology, Analysis, Probability \& Statistics, Applied Mathematics, Discrete Mathematics, and Others.

Conceptual difficulty denotes the knowledge level required for problem solving and is defined with reference to the U.S. education curriculum, spanning six tiers: elementary, middle school, high school, undergraduate, graduate, and advanced research.
Reasoning difficulty characterizes the structural complexity and depth of multi-step reasoning involved in a solution, and is categorized into four levels: shallow, moderate, deep, and extremely hard.
\begin{figure*}[t]
    \centering
    \begin{minipage}{0.45\textwidth}
        \centering
        \includegraphics[width=\linewidth]{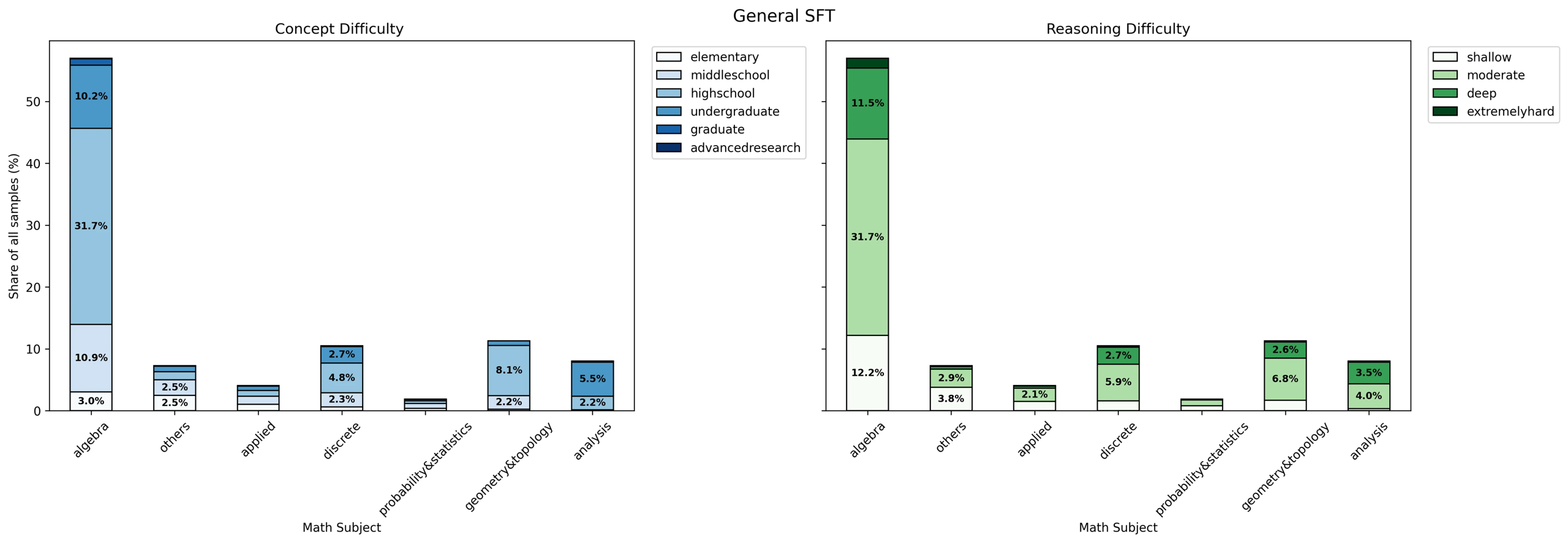}
        {\small (a) Domain$\times$difficulty distribution for general mathematical data.}
    \end{minipage}
    \hfill
    \begin{minipage}{0.45\textwidth}
        \centering
        \includegraphics[width=\linewidth]{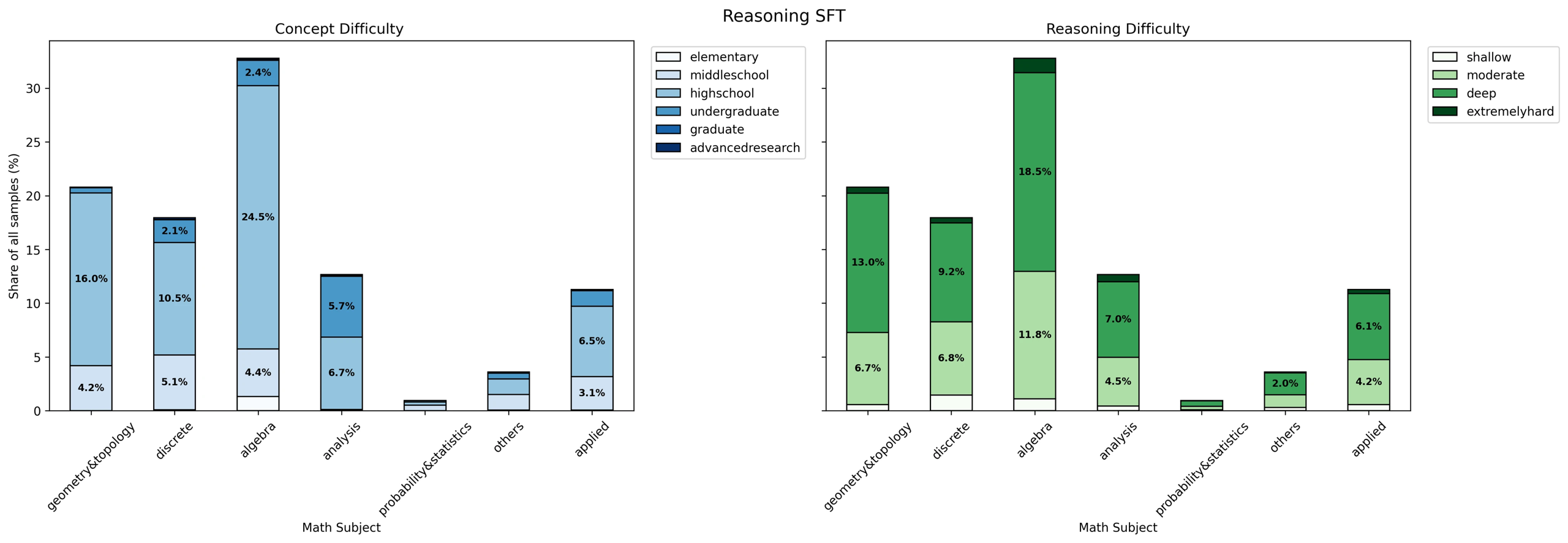}
        {\small (b) Domain$\times$difficulty distribution for mathematical reasoning data.}
    \end{minipage}

    \vspace{0.8em}

    \begin{minipage}{0.45\textwidth}
        \centering
        \includegraphics[width=\linewidth]{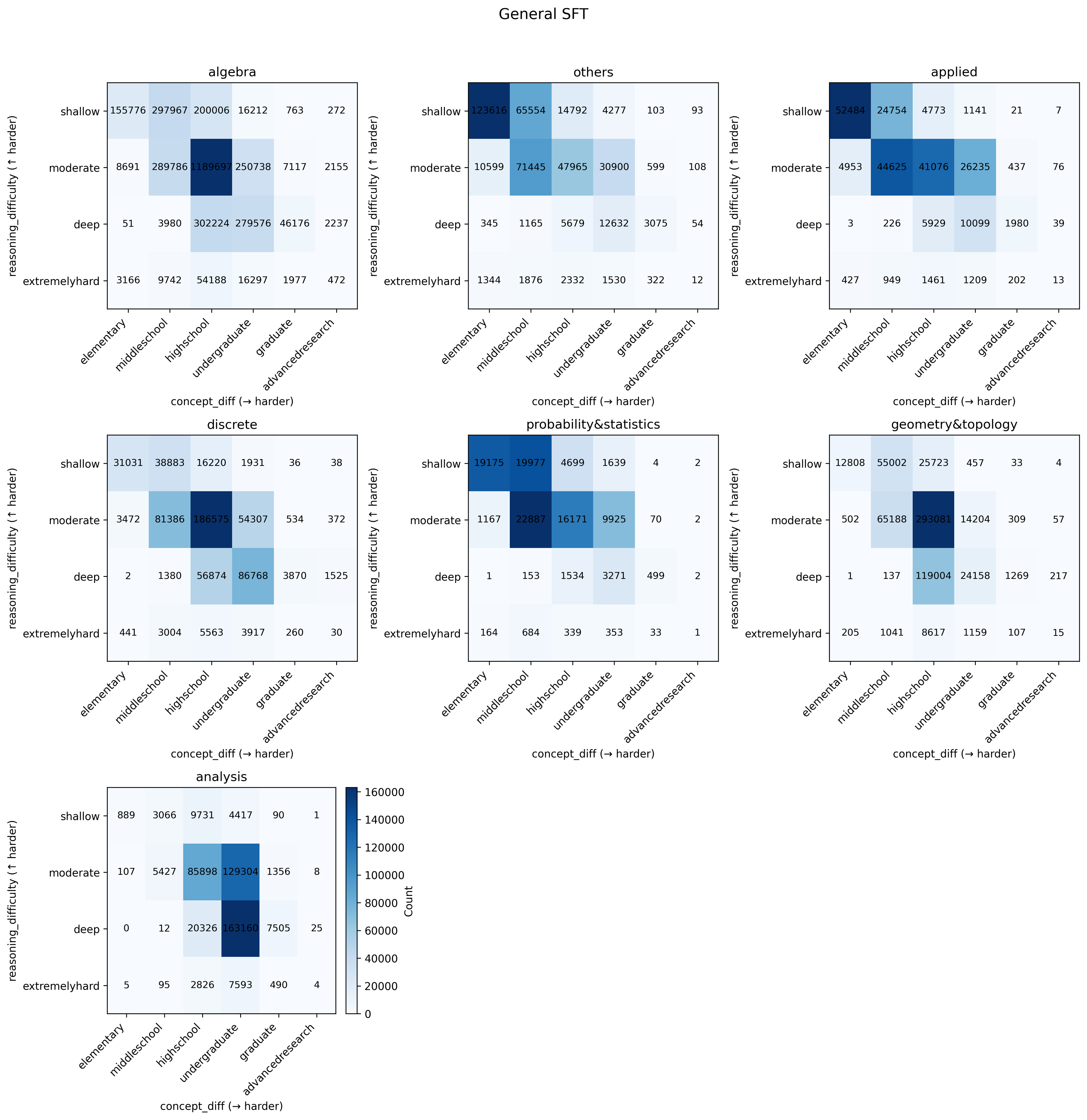}
        {\small (c) Conceptual difficulty$\times$reasoning difficulty distribution for general mathematical data.}
    \end{minipage}
    \hfill
    \begin{minipage}{0.45\textwidth}
        \centering
        \includegraphics[width=\linewidth]{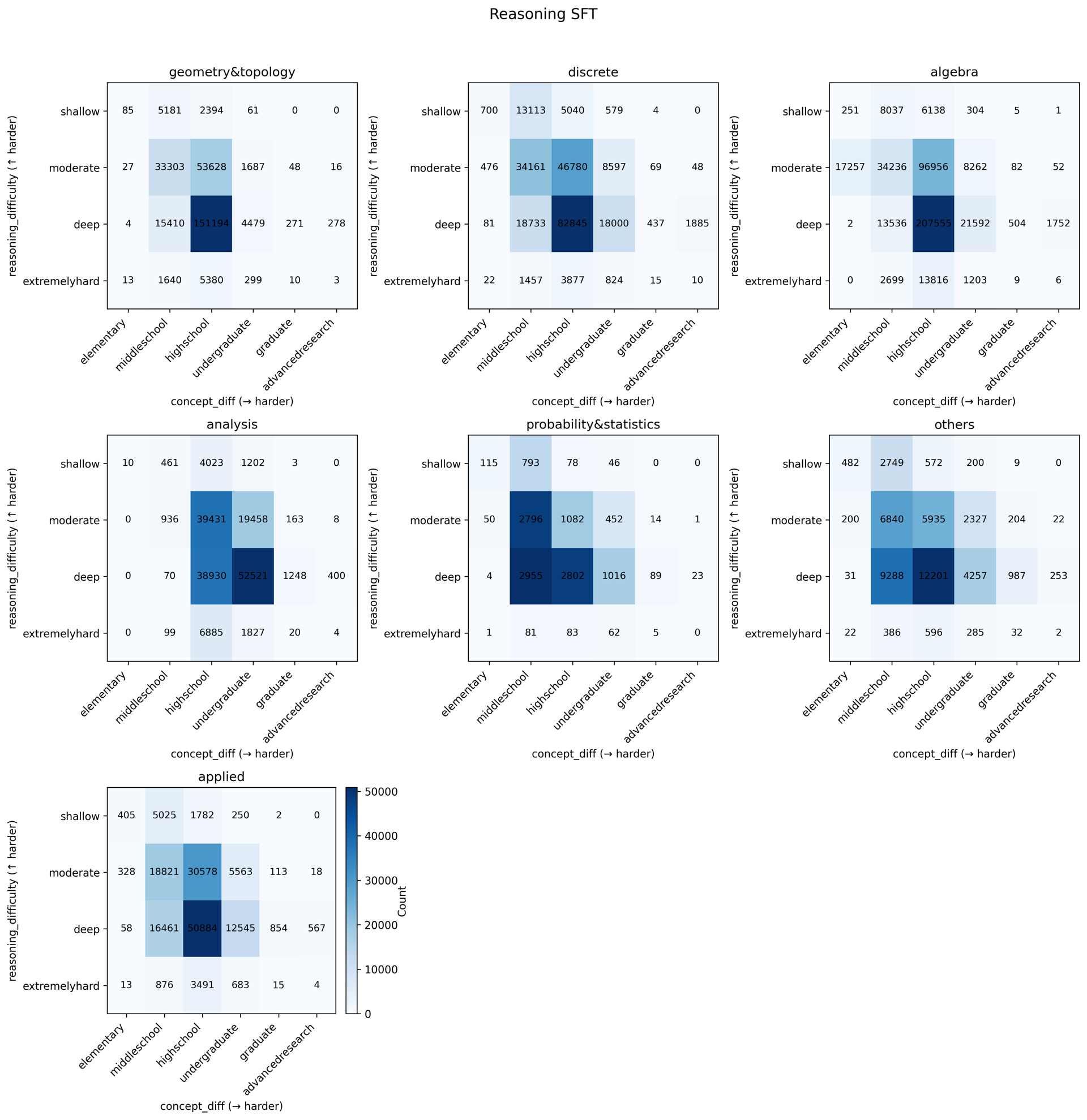}
        {\small (d) Conceptual difficulty$\times$reasoning difficulty distribution for mathematical reasoning data.}
    \end{minipage}

    \caption{Mathematical data distribution according to domain$\times$conceptual difficulty$\times$reasoning difficulty combinations.}
    \label{fig:math_data_distribution}
\end{figure*}

Using the tagging standards defined above, we analyze distributions over the full mathematical corpus as well as each constituent dataset. The results reveal pronounced imbalances in both domain composition and the joint distribution of domain and difficulty. At the corpus level, Algebra and Analysis account for the largest share of samples, whereas Geometry \& Topology, Applied Mathematics, and Discrete Mathematics constitute a relatively smaller portion of the corpus.

When analyzed jointly with difficulty, domain imbalance becomes more pronounced. In Algebra and Analysis, most problems cluster at the elementary–high school conceptual levels and require shallow–moderate reasoning. By contrast, in Geometry \& Topology and Applied Mathematics, problems at the undergraduate level or above and those requiring deep (or higher) reasoning are markedly under-represented. This indicates structural gaps not only in overall domain coverage but also in the domain$\times$difficulty space.

This distribution can induce structural biases beyond mere data scarcity: models may be repeatedly exposed to certain modes of mathematical thinking while receiving insufficient coverage of others, including spatial reasoning, diagram-based reasoning, equation/formula modeling in applied contexts, and combinatorial reasoning over discrete structures. Notably, the commonly observed pattern of early gains followed by a plateau on benchmarks such as GSM8K~\cite{cobbe2021gsm8k} and MATH~\cite{hendrycksmath2021} could plausibly be linked to this domain and reasoning-type imbalance.

To mitigate this, we perform targeted, gap-filling synthesis focused on under-covered regions of the domain$\times$difficulty space. We prioritize domains with limited coverage—Geometry \& Topology, Applied Mathematics, and Discrete Mathematics—as well as problems at undergraduate-or-higher conceptual levels and/or requiring deep-or-higher reasoning. Conditioned on core concepts, we generate problem–solution pairs with controlled reasoning depth, and we convert simple QA-style items into formats that include step-by-step solutions. The resulting synthetic corpus is annotated using the same domain, conceptual difficulty, and reasoning difficulty tagging criteria and is incorporated into our reasoning-data construction.

\subsection{Mi:dm K Response Style Principles}
\label{subsection2-4:data-styleguide}


The quality of model responses depends not only on content accuracy but also on the stability and structural readability of the output format. In particular, for long-form answers, comparison and summarization queries, and procedural guidance requests, the absence of consistent visual conventions can fragment key information, degrading user experience and increasing post-processing overhead~\cite{li2025scar, chen2025mdeval}. Accordingly, to ensure consistent formatting, we define the \emph{Mi:dm K Response Style Principles} and rewrite SFT responses to conform to these principles. The principles comprise 15 elements—covering emphasis, numeric and year notation, indentation, and bullet usage—and are designed to enforce structure at the sentence, paragraph, and document levels.

This form of response rewriting is most effective when stylistic consistency has a more direct impact on user experience than fine-grained task semantics. We therefore define such cases as \emph{general tasks} and restrict style-based rewriting to this category. General tasks comprise open-ended tasks without a fixed answer format, including question answering, text generation, comprehension, and analysis. For these tasks, we prioritize consistency in response length, writing style, persona, and structural organization over strict adherence to task-specific templates. As a result, Mi:dm K 2.5 Pro is trained to maintain a stable tone and structure even in the absence of explicit formatting instructions.

By contrast, style-consistency rewriting can be inappropriate when output formats or constraints are strict—for example, problem-solving data in the code and mathematics domains; simple QA or instruction-following (IF) data with fixed formats; and data involving agent workflows, multi-turn dialogue, safety, or reasoning-intensive tasks. In such cases, over-emphasizing stylistic uniformity can conflict with task-specific requirements. We also exclude data with mixed task types or low classification confidence from rewriting, as enforcing a uniform style may introduce unintended artifacts. Finally, when user instructions explicitly specify output formats or constraints, we prioritize instruction adherence over the style principles. 

In addition, to ensure safety against harmful requests, we separately incorporate a Responsible AI Response Style addressing content-level considerations. For adversarial prompts, the default policy is to refuse, deflect, or provide only a restricted response; however, the appropriate response form—in terms of both content and format—is determined by taking into account the prompt type (i.e., intent) and topic category.

\begin{table}[h]
\centering
\setlength{\tabcolsep}{5pt}
\renewcommand{\arraystretch}{1.1}
\resizebox{\textwidth}{!}{
\begin{tabular}{lccccccc}
\toprule\midrule
\rowcolor{gray!10}
  & \textbf{Character} & \textbf{Line Break(\texttt{\textbackslash n})} & \textbf{Characters/Line} & \textbf{Paragraph} & \textbf{Heading} & \textbf{Bullet} & \textbf{Code Block} \\
\midrule
Before Rewriting
& 742.92
& 7.87
& 227.07
& 4.53
& 0.68
& 1.46
& 0.10 \\

After Rewriting
& 2649.99
& 69.85
& 43.40
& 25.06
& 4.90
& 40.47
& 0.14 \\\hdashline

$\Delta$ (Diff.)
& \makecell{+1907.07\\(256.7\%)}
& \makecell{+61.98\\(787.2\%)}
& \makecell{--183.66\\(--80.9\%)}
& \makecell{+20.53\\(453.1\%)}
& \makecell{+4.22\\(618.6\%)}
& \makecell{+39.01\\(2669.9\%)}
& \makecell{+0.04\\(42.6\%) }\\
\bottomrule
\end{tabular}
}

\caption{Response structure statistics before and after rewriting.}
\label{tab:rewrite_stats}
\end{table}







\begin{figure}[h!]
    \centering
    \begin{minipage}{0.6\linewidth}
        \centering
        \includegraphics[width=\linewidth]{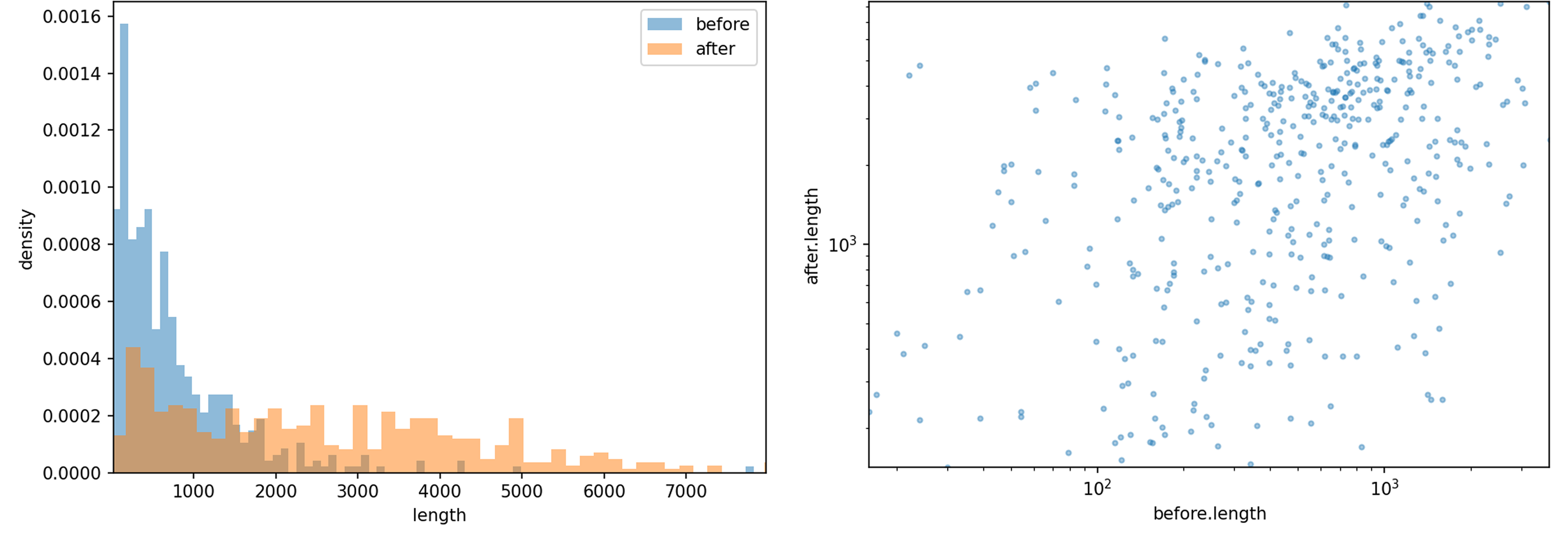}
        \caption{Response length distribution (before vs. after rewriting)}      
        \label{fig:2-6-1}
    \end{minipage}
    \hfill
    \begin{minipage}{0.33\linewidth}
        \centering
        \includegraphics[width=\linewidth]{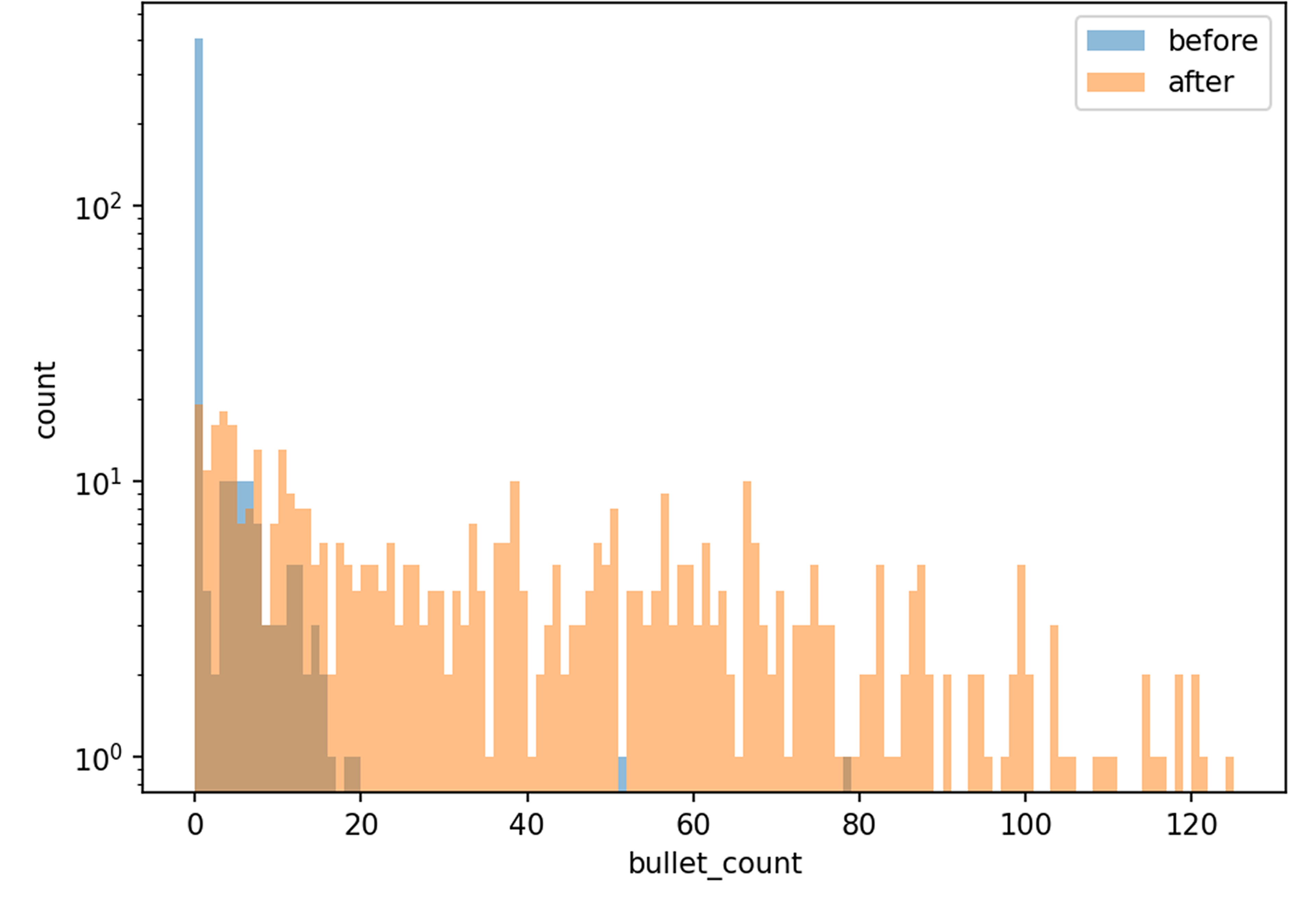}
        \caption{Bullet-count distribution (before vs. after rewriting)}       
        \label{fig:2-6-2}
    \end{minipage}
\end{figure}

\cref{tab:rewrite_stats} compares quantitative metrics and distributional statistics before and after response rewriting. After rewriting, we observe both longer responses and a higher degree of structural organization: the mean response length increases by 256.70\%, reflecting a shift toward more detailed explanations~\cref{fig:2-6-1}. We also observe a substantial increase in bullet usage. As shown in \cref{fig:2-6-2}, the share of responses with {\small\texttt{bullet\_count = 0}} drops from 82.8\% to 3.9\%, indicating that most rewritten responses include bullet-list structure.

These changes reflect not merely longer responses, but a shift toward a more structured presentation in which information is organized into discrete, itemized units. Qualitative analysis further suggests that rewritten responses move beyond short, answer-only statements toward explanation-driven structures that present conditions and context in a more logical, step-by-step manner. As a result, even for general queries with absent or weak instructions, variability in response length, tone, and persona is reduced, indicating improved service-level consistency and more reliable delivery of high-quality responses.

\subsection{Difficulty-Aware Quality Assessment}
\label{subsection2-5:sft-qualityfilter}
\begin{table}[h!]
\centering
\resizebox{\textwidth}{!}{
\begin{tabular}{p{3.2cm} p{5.2cm} p{5.2cm} p{5.2cm}}
\toprule
\multicolumn{1}{c}{\textbf{Assessment Criteria}} &
\multicolumn{1}{c}{\textbf{Definition}} &
\multicolumn{1}{c}{\textbf{\makecell{High Quality\\Data Requirements}}} &
\multicolumn{1}{c}{\textbf{\makecell{Low Quality\\Data Examples}}} \\
\hline\midrule
\rowcolor{gray!10}
\multicolumn{4}{l}{\textbf{\textit{{(A) Quantitative Assessment Criteria for High Quality Data}}}} \\
\midrule

Response Diversity Based on Pre-Training Knowledge      
& Reliable responses and expression diversity based on pre-training knowledge   
& Does not cause catastrophic forgetting and no excessive repetitive patterns are seen
& Repeated similar question answer structures and responses that terminate at a superficial level   \\ 
\hdashline
Capability to Handle High Difficulty Tasks          
& Support performance on high difficulty benchmarks such as GPQA, MATH-500, MMLU-Pro    
& Include accurate reasoning paths without hallucination       
& Citation of non existent facts, statistics, logical leaps, incorrect causal relationships, fictitious API usage \\   
\hdashline
Multilingual Performance Balance        
& Performance balance across supported languages (e.g., Korean, English, and etc.)     
& Natural expressions in each language without awkward translated style      
& Literally translated expressions, English sentence structures for non-english, particle, ending errors \\    
\hdashline
Securing Specialized Capabilities        
& Reasoning, RAG, agentic, and long-context capabilities        
& Include high quality examples appropriate for each scenario     
& Lack of specialized scenario data or appropriate examples \\      
\hline\midrule 
\rowcolor{gray!10}
\multicolumn{4}{l}{\textbf{\textit{(B) Qualitative Assessment Criteria for High Quality Data}}} \\
\midrule
System Prompt Acceptability
& Ability to comprehend and consistently comply with system instructions    
& Compliance with system prompt, prioritization in case of conflict, positional robustness       
& Persona deviation, system constraint violation, role consistency collapse \\       
\hdashline
Instruction Following
& Accurate execution of diverse forms of instructions
& Output format compliance, constraint adherence, and persona consistency
& Format noncompliance, ignoring length limits, and missing requested items \\
\hdashline
New Task Generalization
& Generalization capability for out of tasks (OOT)
& Inclusion of various task types and creative solution cases
& Exclusive generation of template responses, excessive bias toward specific tasks \\
\hdashline
Cultural and Contextual Alignment
& Provide responses that align with culture and context
& Reflect Korean sentiment, comply with the honorific system, recognizes local context
& Western centric examples, cultural misunderstandings, inappropriate honorific usage \\
 \hline\midrule
\rowcolor{gray!10}
\multicolumn{4}{l}{\textbf{\textit{(C) Safety and Consistency Related Assessment Criteria}}} \\   
\midrule
Safety Balance
& Balance between under-refusal and over-refusal
& Rejection of harmful requests while faithful response to legitimate requests
& \textbf{Over}: Rejection of harmless and educational purpose requests \newline
\textbf{Under}: Partial compliance to harmful requests and  vulnerability to jailbreak \\
\hdashline
Honesty \& Transparency
& Awareness of model limitations and transparent communication
& Explicit acknowledgment of uncertainty, hallucination avoidance, and no sycophancy
& Agreement with false premises, failure to point out errors, excessive praise \\
\hdashline
Role Consistency
& Maintaining consistency in persona, role
& Does not show role deviation during conversation
& Persona collapse mid conversation, contradict with system settings \\ 
\hdashline
Response Consistency
& Consistent responses across identical or similar queries
& Stable factual answers and absence of self-contradiction
& Conflicting responses to identical queries, self-contradictions within the same conversation \\
\bottomrule
\end{tabular}
}
\caption{SFT data quality assessment criteria}      
\label{tab:sft_quality_criteria_all}
\end{table}

LLM performance is strongly influenced by the quality of supervised fine-tuning (SFT) data used during alignment, in addition to model architecture and pretraining scale. Lima~\cite{zhou2023lima} shows that strong alignment can be achieved with roughly 1,000 high-quality samples, underscoring the importance of data quality at the SFT stage. In contrast, incorporating low-quality data can lead models to internalize undesirable behaviors—such as hallucinations, IF failures, and inappropriate safety judgments—thereby reducing alignment quality.
In practice, SFT dataset construction typically draws on both public and synthetic data, which are subject to non-trivial quality variance due to their generation and distribution mechanisms. Public datasets are often released in bulk, making fine-grained, per-sample quality auditing difficult and thereby allowing substantial noise to persist. Synthetic data can likewise exhibit issues—including hallucinations, logical inconsistencies, and cultural inappropriateness—stemming from limitations of the generating model~\cite{ji2023survey}. Moreover, prior work provides empirical evidence that heterogeneous data quality can negatively impact model performance~\cite{longpre2023flan}.

To mitigate these issues, we introduce an automated quality assessment framework that identifies and filters low-quality samples from large-scale SFT corpora. We define multi-dimensional quality criteria and perform quality tagging and filtering using an LLM-based data evaluator. By removing samples that are clearly low-quality before training, this framework improves the overall effectiveness of alignment.

\paragraph{Multi-Dimensional Criteria and Interdependencies.}
Systematic assessment of SFT data quality requires a multidimensional perspective rather than a single criterion. Evaluations limited to grammatical correctness or formal completeness are insufficient to anticipate a model’s real-world performance and behavior when trained on such data. We therefore define SFT quality criteria along three axes: quantitative performance, qualitative performance, and helpfulness/safety. The detailed criteria for each axis are summarized in \cref{tab:sft_quality_criteria_all}.

The three quality criteria defined above are not independent; rather, they are interdependent and interact with one another, as summarized in \cref{tab:interaction_relationships}. For instance, strong quantitative performance (e.g., robust reasoning ability) underpins qualitative performance by enabling accurate execution of complex instructions. When qualitative performance is unstable, consistent compliance with safety guidelines can also degrade. Moreover, safety and helpfulness exhibit an inherent trade-off: excessive emphasis on one can undermine the other.

Moreover, low-quality signals often manifest across multiple criteria simultaneously rather than along a single dimension. We therefore assess data quality holistically, considering the balance and interactions among criteria instead of evaluating each criterion in isolation.

\begin{table}[h!]
\centering
\resizebox{\textwidth}{!}{
\begin{tabular}{p{6.5cm} p{10.5cm}}
\toprule\midrule
\rowcolor{gray!10}
\textbf{Interaction} & \textbf{Relationship} \\
\midrule
Quantitative performance $\leftrightarrow$ Qualitative performance
& Strong reasoning capability is the foundation for complex instruction following \\
\hdashline
Qualitative performance $\leftrightarrow$ Safety
& In case of low acceptability of the system prompt, safety guideline compliance also becomes unstable \\
\hdashline
Safety $\leftrightarrow$ Helpfulness
& Excessive emphasis on safety impairs helpfulness, and vice versa \\
\bottomrule
\end{tabular}
}
\caption{Interactions and trade-offs among performance, safety, and helpfulness}
\label{tab:interaction_relationships}
\end{table}
\paragraph{Data Quality Assessment Framework.}
Building on the defined quality criteria and the analyzed problem types, we develop an LLM-based data assessment model to efficiently identify low-quality samples in large-scale SFT corpora. To ensure the reliability of the assessment model, we adopt a human-in-the-loop iterative refinement process. The overall workflow consists of three stages: data sampling, human assessment, and iterative prompt refinement.


First, to improve the generality of the assessment model, we sample an evaluation set in a controlled manner across multiple dimensions. We balance response length by including samples ranging from short to long outputs, and ensure topical diversity by covering task types such as general conversation, coding, reasoning, and creative writing. To further mitigate source bias, we perform stratified sampling across purchased data, public datasets, and internally generated data.

Next, data-quality experts directly evaluate the sampled data. Each evaluation comprises (i) a binary error judgment ({\small\texttt{True/False}}), (ii) error-category annotation based on the criteria in \cref{tab:sft_quality_criteria_all}, and (iii) a written rationale supporting the judgment. This process yields a gold-standard dataset for validating the assessment model.

Finally, we evaluate the LLM-based assessment model against the expert labels and iteratively refine the prompt by analyzing disagreement cases. After applying an initial prompt, we measure accuracy relative to expert judgments, examine false positives and false negatives, and update the prompt accordingly. Repeating this cycle progressively improves agreement with expert assessment.

We design the prompt for the LLM-based data assessment model around three core principles. First, we simplify the decision structure by using a binary validity judgment ({\small\texttt{valid}: \texttt{true/false}}) instead of a complex scoring scheme. Second, we enforce a JSON output format to ensure structured, machine-readable results. Third, when issues are detected, we require the model to include both a high-level issue category and a textual explanation so that the rationale is explicit.
Accordingly, the final prompt outputs {\small\texttt{\{"valid": false, "issues": [\{"category": "<high-level category>", "explanation": "<issue description>"\}]\}}} when issues are present, and {\small\texttt{\{"valid": true, "issues": []\}}} otherwise. This binary, structured design clarifies filtering criteria, improves assessment consistency, and simplifies downstream post-processing and aggregation.

After iterative refinement, the final assessment prompt attains 84.2\% accuracy against human judgments, with 86.8\% precision and 83.6\% recall. To further validate the practical utility of the evaluator, we compare pass rates before and after rewriting samples flagged as low quality. Across 929 cases, the pass rate increases by approximately 6.8 percentage points after rewriting, suggesting that the evaluator can discriminate meaningful quality differences in practice.

Finally, we filter the data using the validated assessment model. Samples with {\small\texttt{valid: false}} that contain critical-category issues are labeled as \emph{Reject}, whereas samples with {\small\texttt{valid: false}} containing only minor-category issues are labeled as \emph{Review}. Samples judged as {\small\texttt{valid: true}} are labeled as \emph{Pass}, and we restrict exclusion from training to \emph{Reject} only. Samples assigned to \emph{Review} undergo additional inspection to determine whether they should be included.
\section{Pre-Training}
\label{section3:pre-training}
In this section, we describe the key design elements for enhancing reasoning capability during the pre-training stage of Mi:dm K 2.5 Pro. Whereas Mi:dm 2.0 Base pre-training primarily targeted general language modeling performance, Mi:dm K 2.5 Pro expands pre-training data composition and training strategy to better meet the reasoning demands of subsequent post-training stages. Specifically, we introduce reasoning-centric continual pre-training (CPT) data expansion, depth upscaling (DuS) for efficient model scaling, and a gradual context-length extension strategy to maintain long-context capability. We focus on the design elements that most clearly distinguish Mi:dm K 2.5 Pro from Mi:dm 2.0.

\subsection{Data Expansion}      
\label{subsection3-1:CPT Data}
To ensure the model with sufficient reasoning capability for subsequent post-training stages (e.g., SFT and RL), we design a CPT data expansion strategy that goes beyond collecting public reasoning datasets by explicitly incorporating reasoning processes. The goal is to introduce reasoning-centric supervision ahead of post-training, so that the model can learn coherent and consistent reasoning trajectories for problem solving in specialized knowledge domains.

\paragraph{Curriculum-Guided Reasoning for STEM Domains.}      
Unlike general commonsense queries, problems in expert domains often presuppose substantial prerequisite knowledge as well as specific theoretical, experimental, or measurement contexts. Moreover, they are frequently posed as compound queries rather than single questions, typically requiring at least 2–3 reasoning steps. Consequently, datasets dominated by simple question–answer pairs are insufficient for developing robust and stable reasoning capability in specialized domains.

Accordingly, we first construct a reasoning knowledge framework grounded in course curriculum. We organize knowledge along subject- and major-level curricula and formalize it as a taxonomy that can be used as input to the data synthesis pipeline. We then expand this taxonomy into concrete training objectives that go beyond enumerating what should be known, explicitly capturing a reasoning perspective by specifying the conditions under which concepts should be applied and the conclusions that should be derived.

Based on these training objectives, we construct data via a staged synthesis pipeline—{knowledge document $\rightarrow$ question $\rightarrow$ reasoning path}—rather than directly augmenting individual items. For each objective, we first compile a knowledge document that captures key definitions, prerequisites, exceptions, and common confounders required for reasoning. We then generate multi-step QA instances with controlled reasoning depth. Each instance comprises an introduction that sets up the problem context and an application step that applies the relevant knowledge to derive a conclusion, and each QA is accompanied by an explicit reasoning path. Boilerplate or insufficient reasoning paths are detected using rule-based checks and strengthened via rewriting, ensuring that coherent reasoning flow is learned consistently during CPT.

\paragraph{Enhancing Procedural Reasoning in Code.}      
To effectively leverage reasoning-oriented code data for training, we go beyond coarse classification by difficulty, language, or correctness. We systematically filter samples that are unsuitable for learning while maintaining a composition that reflects real-world distributions over programming languages, difficulty levels, and task types. Accordingly, during the CPT stage, we apply the code refinement and tagging pipeline described in \cref{subsection2-2:data-code}. The pipeline sequentially performs programming-language identification, primary refinement based on educational suitability and quality criteria, source- and file-level noise removal, AST-based executability filtering, and difficulty and task labeling.
The resulting metadata makes the problem-solving context and procedural complexity of each code sample explicit, providing a basis for broad coverage of procedural reasoning patterns and problem-solving types in the code domain during pre-training.

\paragraph{Diverse Reasoning Coverage in Math.}          
During CPT for mathematics, we prioritize broad exposure to diverse mathematical reasoning structures and concept combinations, rather than focusing solely on answer derivation for individual problems. The data composition follows the domain- and difficulty-based classification principles defined in \cref{subsection2-3:data-math}. Since mathematical difficulty is determined not only by conceptual level but also by the number of reasoning steps required to apply those concepts, we curate a balanced mixture spanning different conceptual levels and reasoning depths. This design encourages the model to learn a wide range of mathematical reasoning patterns without overfitting to specific computation types or problem formats during CPT, providing a foundation for more stable generalization to challenging mathematical reasoning tasks in subsequent post-training.

\subsection{Depth Upscale}      
\label{subsection3-2:DUS}

Depth upscaling (DuS) is a model scaling technique that increases model depth by reusing the parameters of an existing model~\cite{young2024yi, yang2025lesa}. This approach provides an efficient way to expand model capacity and was previously adopted in the training of Mi:dm 2.0~\cite{shin2026mi}. In this work, we again employ DuS to enable rapid scaling toward a substantially larger model, Mi:dm K 2.5 Pro.
For Mi:dm K 2.5 Pro, we systematically evaluate two representative model scaling strategies: cosine similarity based scaling~\cite{young2024yi}, which was employed in Mi:dm 2.0, and layer predictor based scaling~\cite{yang2025lesa}. To analyze their effects under different architectural conditions, we conduct controlled experiments across a range of layer expansion positions.
Layer predictor based scaling constructs newly added layers by predicting their parameters through a dedicated layer predictor. This predictor learns inter-layer parameter relationships from adjacent layers, which are extracted via singular value decomposition. By leveraging these learned relationships, the method enables coherent parameter initialization for deeper architectures while preserving structural continuity across layers.

The existing cosine similarity based scaling requires extensive iterative experimentation to identify an optimal combination of layers that yields strong performance after scaling. In contrast, layer predictor based scaling initializes the expanded layers using pre-predicted parameters, leading to a more stable starting point for learning and substantially reducing the search cost. This advantage is particularly important because the performance of models using DuS strongly depends on the expressiveness of the layers selected for scaling. Accurately identifying layers with high expressiveness directly influences both the initial performance and the training stability of the scaled model.

\begin{table}[h!]
\centering
\scriptsize

\resizebox{0.8\textwidth}{!}{
\begin{tabular}{l c c c c}
\hline
 & \textbf{MMLU} & \textbf{KMMLU} & \textbf{GSM8K} & \textbf{MBPP} \\
 & (5-shot) & (5-shot) & (5-shot) & (3-shot) \\
\hline
Baseline
& 0.772 & 0.521 & 0.804 & 0.646 \\ \hdashline
+ Random layer based DuS
& 0.758 & 0.516 & 0.667 & 0.574 \\
+ Cosine similarity based DuS 
& 0.767 & 0.514 & 0.782 & 0.624 \\
+ Layer predictor based DuS
& \textbf{0.804} & \textbf{0.572} & \textbf{0.805} & \textbf{0.648} \\
\hline
\end{tabular}
}
\caption{Model performance before and after DuS across different methods.}
       
\label{tab:dus_ablation}
\end{table}

\cref{tab:dus_ablation} compares the initial performance of each model before DuS and immediately after DuS is applied. The performance after DuS refers to a model that undergoes only structural expansion without any additional training. When random layer based DuS~\cite{kim2024solar}, which selects arbitrary layers for depth scaling, and cosine similarity based scaling are applied, model performance generally degrades compared to the baseline before DuS. In contrast, DuS with a layer predictor results in much smaller performance drops and, in some metrics, even leads to performance improvements. Furthermore, after additional training following DuS, we observe consistent improvements across benchmarks, demonstrating that stable performance gains are achievable even after depth expansion. Based on these findings, we adopt the layer predictor method as our DuS scaling strategy.

\begin{table}[h!]
\centering
\setlength{\tabcolsep}{6pt}
\renewcommand{\arraystretch}{1.2}
\resizebox{\textwidth}{!}{
\begin{tabular}{l l c c c c c c}
\toprule
\multirow{2}{*}{\textbf{Model}} & \multirow{2}{*}{\textbf{Training Phase}}
& \textbf{MMLU} & \textbf{MMLU-Pro} & \textbf{KMMLU} & \textbf{GSM8K} & \textbf{MBPP} \\
& & (5-shot) & (5-shot) & (5-shot) & (5-shot) & (3-shot) \\
\midrule
CPT Stage 0 
& After alignment 

& 80.28 & 51.70 & 52.90 & 79.45 & 59.20 \\
CPT Stage 1 
& After training replay data

& 79.11 & 53.21 & 61.88 & 87.57 & 67.40 \\
CPT Stage 1 Merged 
& After model merge 

& 81.11 & \textbf{60.05} & 63.95 & 86.66 & 70.20 \\
CPT Stage 2 
& After focused STEM training

& \textbf{83.11} & 59.84 & \textbf{67.85} & \textbf{89.54} & \textbf{71.20} \\
\bottomrule
\end{tabular}
}
\caption{Comparison of model performance across CPT training stages}            
\label{tab:cpt_stage_results}
\end{table}

After scaling the model with the layer predictor method, we conduct CPT in three stages, starting with a stabilization phase that consolidates the expanded layer structure introduced by DuS. \cref{tab:cpt_stage_results} reports the changes in model performance across each CPT stage.
To ensure stable training of the scaled model, we first conduct alignment training before full scale CPT. We refer to this as CPT stage 0. CPT stage 1 then focuses on stabilizing overall model performance after scaling, using a data composition of 29\% English, 17\% Korean, 50\% STEM, and 9\% multilingual data. After stage 1, we apply a model merging technique using checkpoints saved at regular intervals~\cite{tao2024unlocking}. Compared to the final checkpoint from stage 1, the merged model shows improved performance on most evaluation metrics, with the exception of GSM8K~\cite{cobbe2021gsm8k}.
In CPT stage 2, we increase the proportion of STEM data relative to the previous stage in order to further enhance overall performance and strengthen reasoning capability.
\subsection{Long Context Expansion Strategy}         
\label{subsection3-3:pre-training}
To enhance the long context processing capability of Mi:dm K 2.5 Pro, we apply a progressive context length expansion strategy~\cite{yang2025qwen2} during the final stage of pre-training. Across three expansion stages, we gradually increase the context length from 4,096 tokens to 32,768 tokens, 65,536 tokens, and finally 131,072 tokens, enabling stable support for inputs up to 128k tokens.
Throughout this process, we aim to preserve performance on both general task benchmarks and long context benchmarks. In addition, to ensure stable compatibility with the vLLM~\cite{kwon2023efficient} inference framework used during the reinforcement learning phase of post-training, we adopt the YaRN mechanism~\cite{pengyarn}.

In the long context expansion stage, we compose 80\% of the total training data with samples designed to elicit the reasoning capabilities required in the subsequent post-training stage. Specifically, we synthesize and use reasoning data such as mathematics and code, along with various types of long question answering data, including multi-document comprehension and information aggregation over long contexts.

We determine the detailed data mixture ratio using RegMix~\cite{liuregmix}, which recursively estimates the optimal proportions based on the performance of a proxy model. Compared to heuristic approaches, RegMix provides an automated strategy for data mixing in environments with diverse domains. We allocate a portion of the training mixture to replay data from previous stages to mitigate potential catastrophic forgetting during long-context expansion.~\cite{yang2025qwen2, tao2024unlocking, taocan}.

To verify improvements in long context processing capability, we conduct evaluations at each context expansion stage using Ruler~\cite{hsieh2024ruler}. In addition, we use MMLU-Pro~\cite{wang2024mmlu} and GPQA-Diamond~\cite{rein2024gpqa} to monitor potential degradation in general performance that may arise from long context training.

\begin{table}[t]
\centering
\small
\setlength{\tabcolsep}{6pt}
\renewcommand{\arraystretch}{1.2}
\resizebox{\textwidth}{!}{
\begin{tabular}{lccccc cccc}
\toprule
\multirow{2}{*}{\textbf{Model}} 
& \multirow{2}{*}{\makecell{\textbf{Training}\\\textbf{Length}}} 
& \multirow{2}{*}{\makecell{\textbf{Training}\\\textbf{Step}}}
& \multirow{2}{*}{\makecell{\textbf{MMLU-Pro}\\(5-shot, CoT)} }
& \multirow{2}{*}{\makecell{\textbf{GPQA-D}\\(5-shot, CoT)}}
& \multicolumn{5}{c}{\textbf{RULER}} \\
\cmidrule(lr){6-10}
& & & & 

& {4K} 
& {8K} 
& {16K} 
& {32K} & {Avg.} \\
\midrule

Midm 2.5 stage-2
& 4K 
& -- 
& \textbf{59.84} 
& 35.30 

& 89.90 
& 79.26 
& 0.10 
& 0.00 & 42.32 \\

Midm 2.5 stage 3-1
& 32K 
& 1,000 
& 59.76 
& 36.90 

& 89.94 
& 88.01 
& 84.68 
& 65.11 & 81.94 \\

Midm 2.5 stage 3-2
& 64K 
& 2,000 
& 58.98 
& 41.90 

& \textbf{94.02} 
& \textbf{93.08} 
& 87.02 
& 73.48 & 86.90 \\

Midm 2.5 stage 3-3
& 128K 
& 200 
& 59.17 
& \textbf{44.44} 

& 91.95 
& 91.71 
& \textbf{88.05} 
& \textbf{83.43} & \textbf{88.79} \\

\bottomrule
\end{tabular}
}
\caption{Performance comparison across long-context training stages.}
\label{tab:long_context_results}

\end{table}
As shown in \cref{tab:long_context_results}, long-context processing capability improves substantially as the context window used in training is extended. In particular, the model extended to 128k demonstrates improved performance on GPQA-Diamond relative to its performance prior to long-context training. These results suggest that potential degradation of existing capabilities during long-context training can be effectively mitigated through an appropriate data-mixing strategy.
\section{Post-Training}
\label{section4:post-training}
This section describes the post-training process of Mi:dm K 2.5 Pro. We describe the supervised fine-tuning (SFT), model merging, reinforcement learning (RL) pipeline that we design to enhance reasoning performance. Fusion SFT strengthens non-reasoning tasks and improves training stability via controlled mixing of reasoning vs. non-reasoning data and within-batch composition constraints. Fusion RL refines reward design to enhance steering, establish model identity, and reduce tool hallucinations, while improving efficiency through format/language/tool constraints and asynchronous LLM-as-a-Judge–based reward computation. Furthermore, we discuss the difficulty control and reward design components for improving training stability and efficiency during the reinforcement learning stage.

\subsection{Data Expansion}
\label{subsec:post_data}
Mi:dm K 2.5 Pro expands its data composition to strengthen three key capabilities: Korean-specific vulnerability mitigation, long conversation context processing, and tool use in agentic environments. To achieve this, we incorporate error patterns and interaction requirements that have been repeatedly observed in real-world settings. This section describes the design rationale and construction methodology of the Korean-centric SFT data, multi-turn conversation data, and multi-turn tool use datasets.

\paragraph{Strengthening Korean Reliability.}
\label{subsubsection2-5:korean_centric_sft_data}
In Korean-language contexts, model limitations extend beyond translation quality and become most apparent when social and cultural context, institutional norms, Korean-specific linguistic processing, and user writing conventions interact. In practice, the model often misrepresents social contexts, such as appropriate use of honorifics, references to national symbols, and the tone required in official documents~\cite{lee2024kornat, kim2024click}. The model also frequently makes errors on queries that require alphabetic, syllable, and word-level analysis~\cite{cho2025thunder}.
In addition, it produces inconsistent output structures for checklist, procedural, and summary queries. It also struggles to properly handle colloquial expressions, including memes and slang, as well as constraints specific to the Korean language~\cite{park2024pragmatic}.

Accordingly, during the data construction, we categorize these vulnerabilities into five types and design strategies corresponding to each type. Specifically, we focus on strengthening the coherence of Korean knowledge and social and institutional norms, systematically improving the Korean native unit processing capabilities, while simultaneously ensuring output format consistency, colloquial and creative domain adaptability, and stable constraint condition compliance. 

First, we focus on improving how the model reflects the Korean social and cultural context and institutional norms. Existing models tend to overgeneralize Korean honorifics, which vary according to relationship, generation, and familiarity, or fail to adequately capture Korean conventions in queries about national symbols and official documents. To mitigate this, we treat honorifics not merely as lexical knowledge but as expressions grounded in usage context and social nuance, enabling the model to generalize appropriate honorific choices in similar situations.
Furthermore, for frequently asked Korean language queries such as administrative districts, public holidays, and currency units, we encourage the model to produce relational knowledge rather than simple lists. Responses include inclusion and boundary relationships, as well as relevant legal and institutional foundations. In addition, for queries related to major KT services, we explicitly define the scope of information under possible and impossible conditions, following conservative description principles to reduce exaggeration and hallucination.

Next, we tackle recurring issues in segment recognition for queries that require analysis or calculation at the consonant, syllable, and word levels. Due to the compositional nature of Korean characters, models often make errors in tasks such as consonant separation, syllable-level counting, and string validation, even when the overall response appears fluent. To improve this, we design response formats that explicitly specify the units used for calculation and analysis, rather than presenting only the final result. This approach frames unit recognition as string-level operations and rule-based reasoning rather than as isolated factual recall, laying the groundwork for systematically reducing errors in spelling, search, counting, and text analysis queries.

Additionally, we improve the naturalness of Korean expression and the robustness of creative and colloquial interaction. For creative tasks such as social media posts and slogan writing, we prioritize context-appropriate style and length for purposes like promotion, encouragement, and campaigns, while avoiding excessive ornamentation or abstraction. Our goal is to emphasize rhythm, emotional impact, and concise messaging rather than maximizing informational coverage.
For colloquial interactions such as trending memes, balance games, and dad jokes, we focus on conversational pacing, shared context, and restrained responses instead of reproducing slang or memes verbatim. This design supports lightweight, natural interactions for Korean users and helps reduce quality degradation caused by unnecessarily verbose explanations.

Lastly, we strengthen reliable adherence to compound constraints that frequently appear in Korean user requests. Under negative constraints such as “do not include A but perform B,” existing models often violate prohibitions even when the response is semantically appropriate. To address this, we treat compliance with prohibitions as a core evaluation criterion and design the generation process to avoid prohibited content naturally. This goes beyond safety policy compliance and improves adherence to real-world constraints, enabling more stable responses.

This approach decomposes Korean language performance into four dimensions rather than relying on a single fluency metric: social context coherence, linguistic unit processing accuracy, naturalness of expression and interaction, and constraint compliance. Each dimension is grounded in recurring real-world failure patterns and aims to more systematically capture the linguistic, social, and formal requirements of Korean language use.

\paragraph{Multi-Turn Dataset Construction.}
\label{subsection2-6:data}

User utterances in real-world settings rarely encode all requirements within a single-turn; instead, intents are often refined or changed as the conversation unfolds. Users may deepen a topic, pivot to a different one, or refer back to earlier turns, yielding nonlinear interaction patterns~\cite{rebedea2024canttalkaboutthis}. Handling such contexts requires capabilities beyond basic multi-turn dialogue, including sustained tracking of conversational state and appropriate responses to evolving and compound situations~\cite{kwan2024mt, bai2024mt}. Accordingly, we synthesize a high-quality Korean multi-turn conversation dataset with three key characteristics: dynamic context management, complex task execution, and diversity and usability assurance. We define the components of high-quality conversation synthesis and employ a pipeline that decouples scenario design from utterance generation based on these components.


We define high quality multi-turn conversations along three dimensions: interaction structure, topic and task, and persona. The interaction structure dimension covers not only context-preserving conversations but also real-world conversational phenomena such as topic shifts, recall of prior information, requests for error correction, and abrupt off-topic remarks. The topic and task dimension captures settings in which multiple topics and tasks are interleaved. To reflect real-world service environments, we establish a dual taxonomy that distinguishes task-oriented elements from conversational elements, incorporating both dimensions into scenario design. The persona dimension ensures diversity in synthetic data by sampling personas from a large profile pool~\cite{ge2024scaling}. Beyond simple role labels, each persona is specified as a combination of speaking style, domain expertise, and instruction/command style, promoting variation in utterance length, lexical choice, and reasoning progression even within the same topic or task~\cite{deitke2025molmo}. This enables the model to learn a broad range of user characteristics and interaction styles without overfitting to specific user types or conversational patterns.


Based on these component definitions, we adopt a two-stage generation method that first designs the scenario forming the skeleton of the conversation, then generates dialogue transcripts based on this scenario. This decoupling of conversation-level structural design from turn-level utterance generation ensures both contextual coherence and structural completeness in multi-turn dialogues. 

In stage 1 (Scenario Design), we take a user persona, topic and task lists, and multi-turn conversation patterns as input to design the overall conversational flow—including conversation introduction, topic transitions, and specific request sequences. To mitigate generation bias toward persona-aligned topics or specialized domains, we generate multiple diverse scenarios per persona and sample strategically. 

In stage 2 (Dialogue Transcript Generation), we synthesize multi-turn dialogues that preserve the scenario's intended topics and task objectives while allowing turn-level linguistic flexibility. We reflect persona-specific speaking styles in actual utterances and incorporate instruction following constraints (e.g., "summarize in 3 sentences") to create precise, high-quality user queries. To capture authentic multi-turn dynamics, we prompt the model to use pronominal and demonstrative references to prior turns, and include varied interaction patterns such as error correction, refinement requests, and follow-up questions. By targeting diverse conversation lengths, the dataset covers a broad range of real-world behaviors from brief exchanges to extended dialogues.


\paragraph{Multi-Turn Tool-Use Dataset Construction.}            
\label{subsection2-7:data}
For LLMs to evolve into agents capable of complex decision-making, tool-use capability—the ability to acquire information and perform actions via external tools—is essential. However, single-turn tool call datasets have limitations in that they focus on the accuracy of generating arguments that conform to API specifications rather than flexible interactions based on conversational context. In real-world agentic environments, agents must handle richer interaction patterns, such as requesting missing information when user intent is underspecified, or recognizing when a request is not solvable with the available tools and declining appropriately~\cite{wang2024mtu, shimtooldial, lee2024functionchat}. In Korean-language environments in particular, tool-use datasets that explicitly account for conversational context remain scarce. To address this gap, we construct multi-turn training datasets in both Korean and English to enable comprehensive tool-use behaviors. We apply a stepwise pipeline consisting of tool and scenario design, multi-turn dialogue synthesis, and diversification of system prompts.

To support general-purpose agent use, we select approximately 50 domains including telecommunications, finance, healthcare, and public services—and design domain-specific virtual tools for each. For a subset of domains, we assume a dual-control environment in which both the agent and the user can invoke tools directly, and define user-facing tools accordingly~\cite{yao2024tau}. All tool specifications strictly adhere to the Model Context Protocol (MCP) JSON Schema standard, enabling seamless integration with real APIs and open-source tools~\cite{hou2025model}. Beyond basic function names and parameter types, we provide detailed descriptions for each parameter, including sample values and expected output formats. This enables the model to deeply understand tool capabilities and limitations, ensuring consistent tool calls during multi-turn dialogue generation.

We design tool-based conversation scenarios to mirror realistic tool usage patterns from real-world user interactions by combining three elements: conversational topics and contexts, user personas, and tool sets. Each scenario is structured around realistic contexts where tools can provide clear value to users. We vary conversational style and interaction complexity according to user personas (e.g., age, occupation) to ensure diverse dialogue patterns. Furthermore, by providing tool sets of 1–8 tools—mixing relevant tools with intentional distractors—the model learns to accurately interpret user intent, select only necessary tools, and avoid erroneous calls.

Based on the previously defined scenarios and tool specifications, we synthesize multi-turn conversations featuring organic user-system interactions. To reflect complex decision-making patterns in real agentic environments, we categorize system response patterns into four types: tool execution and result interpretation, proactive information requests, general responses, and rejection of out-of-scope requests~\cite{shimtooldial, lee2024functionchat, ross2025when2call}. We design prompts so these types naturally interweave within conversations, and adjust sampling ratios in training data to avoid bias toward any specific response type.

Different open-source platforms and inference engines (e.g., LMStudio~\cite{lmstudio}, OpenWebUI~\cite{lmstudio}, Ollama~\cite{team2024ollama}) impose distinct tool call specifications. To ensure immediate applicability across diverse ecosystems, we diversify system prompts during training to maximize instruction following capabilities. We establish a two-component pipeline: First, we conduct detailed analysis of platform-specific tool call prompts to define core and optional components, which we dynamically recombine to generate diverse prompt variants. Second, while we standardize the JSON format for tool calls, we vary the enclosing tag formats, strengthening the model's ability to immediately adapt to system-prompt rules in zero-shot settings.

\begin{figure}[h!]
  \centering
  \includegraphics[width=0.85\linewidth]{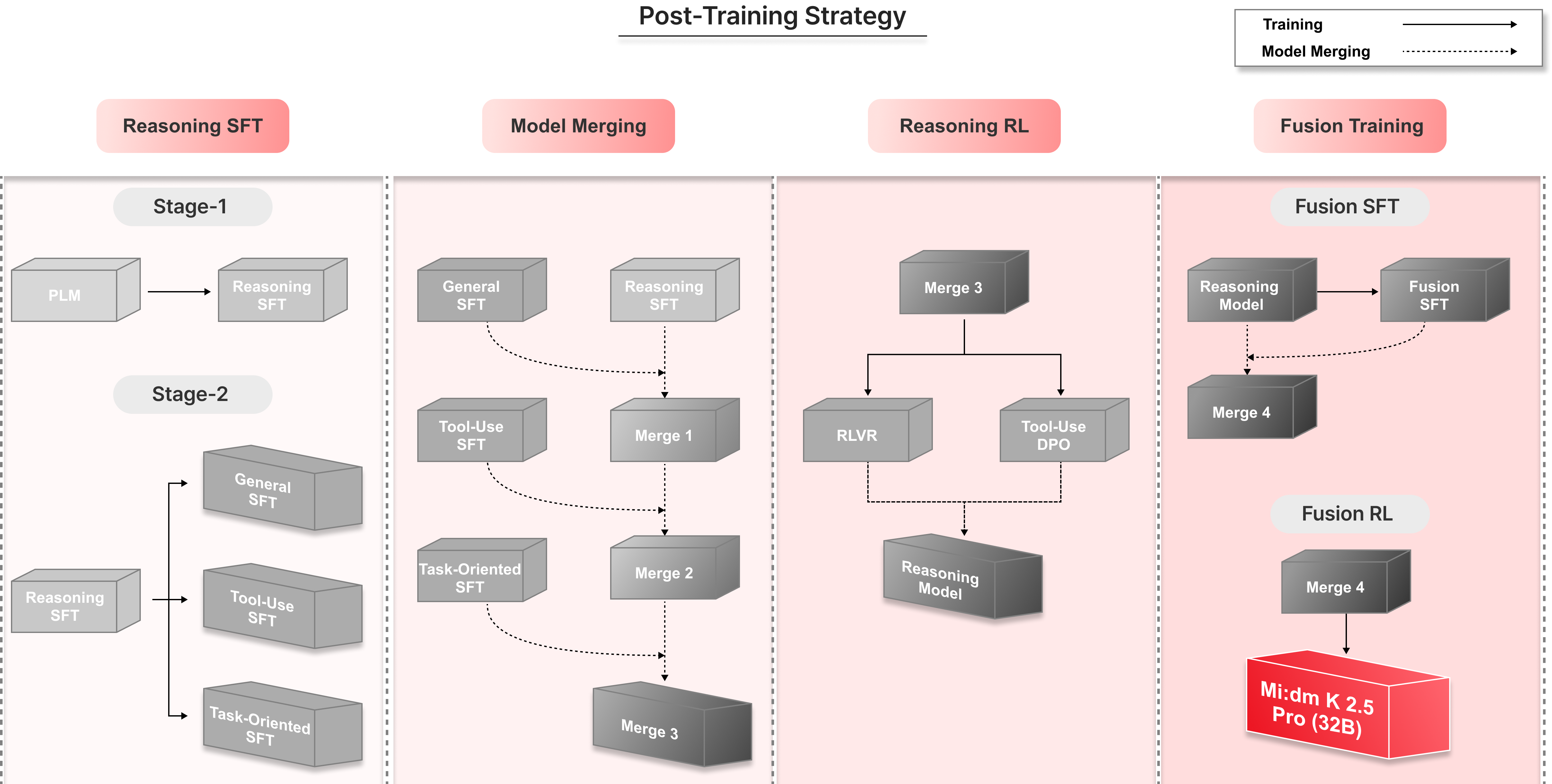}
  \caption{Overview of the post-training pipeline for Mi:dm K 2.5 Pro}
   \label{fig:post-training}
\end{figure}

\subsection{Post-training Design for Reasoning Model}
\label{subsection4-4:post-training}
To enhance the reasoning capability of Mi:dm K 2.5 Pro, we propose a reasoning-specialized post-training strategy that combines reasoning-focused SFT, model merging, and reinforcement learning (RL). \cref{fig:post-training} provides an overview of the full training pipeline.
In the SFT stage, we leverage the model’s extended long-context capability to train step-by-step reasoning traces for complex tasks such as mathematics and coding. This enables the model to refine relevant knowledge, explore diverse reasoning traces, and derive correct answers. We then merge domain-specialized models to construct a general-purpose reasoning model that avoids domain bias. Finally, in the RL stage, we apply verifiable reward-based training to further enhance instruction following, general reasoning, and problem-solving accuracy.
\paragraph{Tokenizer and Chat Template.}
\label{subsection4-1:Tokenizer and Chat Template}

Mi:dm K 2.5 Pro jointly designs the input representation and reasoning interaction structure to enhance reasoning stability and mathematical consistency for numerical data. First, in terms of tokenization, we adopt a 1-digit unit tokenization strategy instead of the 3-digit grouping method used in Mi:dm 2.0. Existing multi-digit number based tokenization can cause issues in which token boundaries distort numerical relationships in numbers with many digits or lead to misrecognition of digit information during computational processes~\cite{singh2024tokenization}. By separating each digit into individual tokens, the model can more precisely learn the numerical context required in arithmetic operations, digit based comparisons, scientific notation, etc., thereby improving the overall consistency and interpretability of numerical reasoning.

\begin{figure}[h!]
  \centering
  \includegraphics[width=0.7\linewidth]{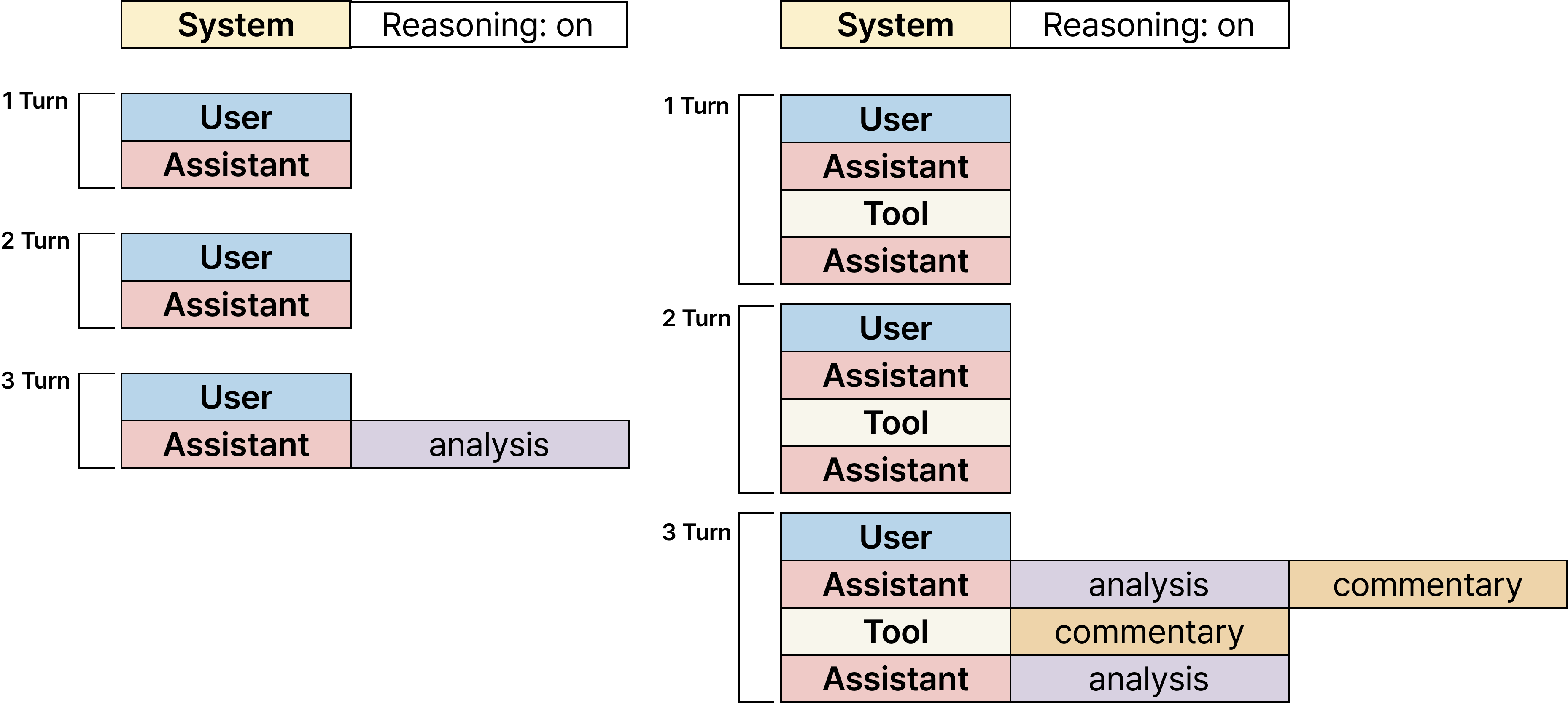}
   \caption{Harmony-based chat template design for efficient SFT training process. (a) Multi-turn reasoning with truncated reasoning trace. (b) Tool-augmented multi-turn interactions retaining only last reasoning trace.
}
   \label{fig:4-2-chat}
\end{figure}

Additionally, during the SFT training process, we use a chat template based on harmony chat format~\cite{agarwal2025gpt}, but apply some modifications considering training efficiency and reasoning control. We adapt this format with reasoning and agent training in mind, and explicitly separate internal reasoning traces from externally visible outputs using channel tokens (\texttt{analysis}, \texttt{commentary}, \texttt{final}) with distinct roles. In Mi:dm K 2.5 Pro, we control reasoning modes via an explicit system-prompt block to improve reasoning controllability and multi-turn training efficiency. During multi-turn reasoning training, we retain only the last reasoning trace associated with the final response or a single tool call, and discard earlier reasoning history to maximize effective context usage. This prevents context waste and reduces noise from accumulated, irrelevant reasoning. \cref{fig:4-2-chat} illustrates an example application of the Harmony Chat Format under these settings.

\begin{table}[h!]
\centering
\scriptsize
\setlength{\tabcolsep}{4pt}
\renewcommand{\arraystretch}{1.05}
\resizebox{\textwidth}{!}{
\begin{tabular}{l c c  c c c  c}
\toprule
\multirow{2}{*}{\textbf{Domain}}
& \multicolumn{2}{c}{\textbf{Reasoning SFT (Stage-1)}} 
& \multicolumn{3}{c}{\textbf{Task-Oriented Training}} 
& \textbf{Reasoning RL} \\
\cmidrule(lr){2-3}
\cmidrule(lr){4-6}
\cmidrule(lr){7-7}
& {Packing Comp.}
& {Ratio}
& {Continual Model}
& {STEM Model}
& {Agent Model}
& {Ratio} \\
\midrule
STEM   & 94\% & 17\% & 17\% & 50\% & --   & 21\% \\
Math   & 85\% & 48\% & 48\% & 27\% & --   & 32\% \\
Code   & 89\% & 34\% & 34\% & 23\% & 57\% & 28\% \\
Agent  & --   & --   & --   & --   & 43\% & --   \\
Instruction Following & -- & -- & -- & -- & -- & 12\% \\
Structured Output     & -- & -- & -- & -- & -- & 12\% \\
\bottomrule
\end{tabular}
}
\caption {Domain mixture ratios across post-training stages and packing efficiency for Reasoning SFT}
\label{tab:reasoning_data_mix}
\end{table}

\paragraph{Reasoning SFT Training Strategy.}
\label{subsubsec: Reasoning SFT}
In the SFT stage, we focus on learning high-quality reasoning traces that leverage the model’s existing knowledge and exploration capacity. To this end, we collect diverse reasoning trace data using multiple models. However, errors in reasoning not only degrade downstream reasoning performance but also increase the likelihood of generating unnecessary tokens during the subsequent RL stage, reducing training efficiency~\cite{guo2025deepseek, matsutani2025rl}. To mitigate these issues, we apply strict post-collection filtering. Specifically, we remove samples exhibiting redundant repetition within reasoning steps, excessive line breaks, and language mixing in final responses. Through this process, we retain only verified, high-quality data as the final training set. We then apply a packing strategy to improve training efficiency for long-form reasoning data and perform task-oriented SFT to progressively incorporate task-specific characteristics while preserving general reasoning capability.

\begin{figure*}[h!]
    \centering
    \begin{minipage}{0.47\textwidth}
        \centering
        \includegraphics[width=\linewidth]{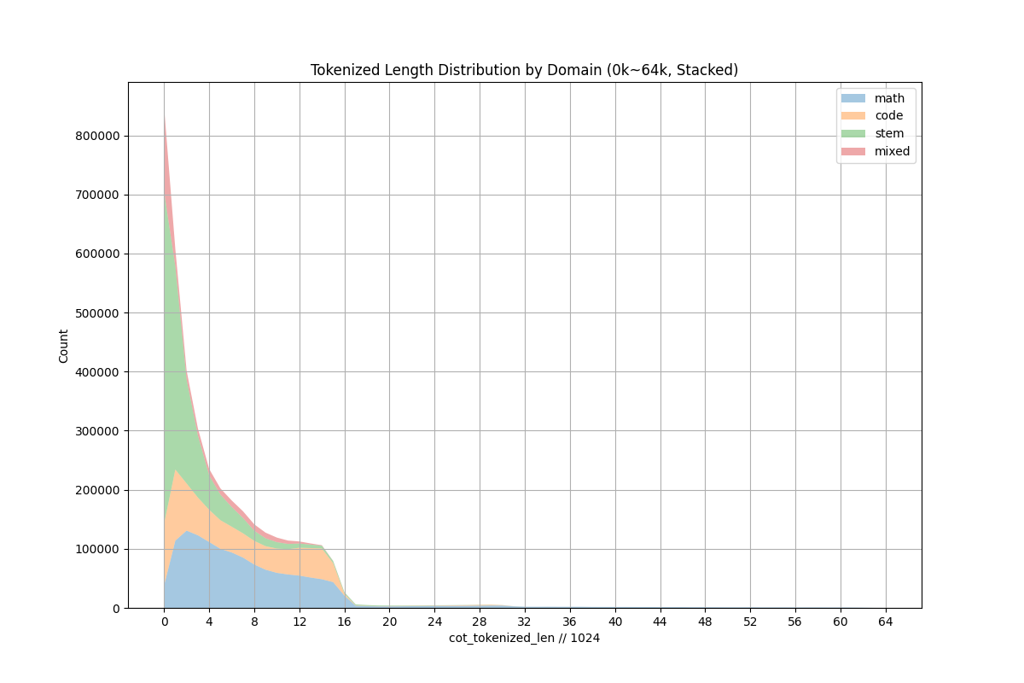}
        {\small (a) Overall length distribution}
    \end{minipage}
    \hfill
    \begin{minipage}{0.47\textwidth}
        \centering
        \includegraphics[width=\linewidth]{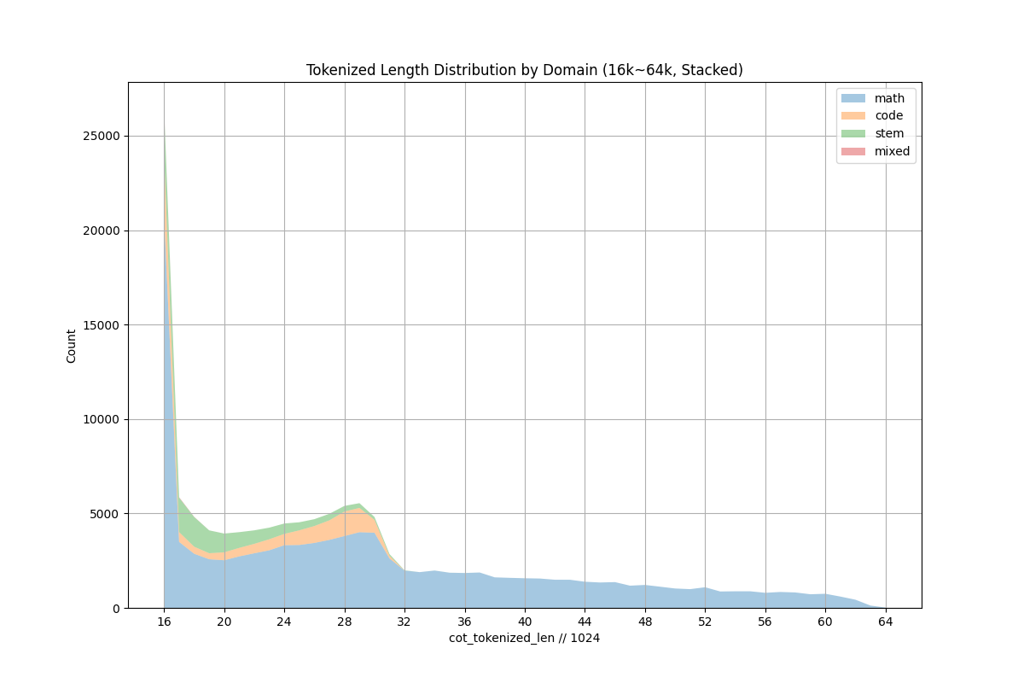}
        {\small (b) Zoomed-in view of the ≥16k length distribution}
    \end{minipage}
    \caption{Response length distribution of collected SFT training data by domain} \label{fig:reasoning_trace_length}
\end{figure*}


The collected reasoning traces exhibit substantial length variability, with some reaching up to 64k tokens. Without packing, padding can dominate a batch—often exceeding the number of effective training tokens—and severely reduce training efficiency. \cref{fig:reasoning_trace_length}(a) shows the response-length distribution of the SFT data by domain; over 80\% of the samples fall within the 0–4k range. We adopt Best bin fit Packing~\cite{ding2024fewertruncationsimprovelanguage} with a 64k maximum sequence length, packing multiple samples into a single sequence. This yields an effective packing efficiency (compression) of 90\% and substantially reduces wall-clock training time. In addition, because training can become unstable when a batch is dominated by a single task type, we apply a blending strategy that samples instances per batch in proportion to the global task mixture.


Reasoning SFT is divided in two stages: the first stage establishes general reasoning capability, and the second stage improves task-specific performance on top of this foundation. Mi:dm K 2.5 Pro aims to preserve both generality and specialization by training a single base model and then deriving multiple task-oriented variants via model merging. To support stable training across stages and improve merging efficiency, we use the Warmup–Stable–Decay (WSD) learning-rate schedule~\cite{hu2024minicpm} for each stage. In the first stage, we heavily emphasize mathematics, code, and STEM data; \cref{tab:reasoning_data_mix} reports sample counts, packing statistics, and mixing ratios. In the subsequent task-oriented SFT stage, we run three additional SFT tracks—continual, STEM-oriented, and tool-oriented—and \cref{tab:reasoning_data_mix} summarizes the corresponding mixture ratios. Finally, we merge the resulting models sequentially, following the procedure in \cref{fig:2-6-2}.

\paragraph{Model Merging.}          
\label{subsection4-5:post-training}

\begin{table*}[h!]
\centering
\small
\setlength{\tabcolsep}{5pt}
\renewcommand{\arraystretch}{1.15}
\resizebox{\textwidth}{!}{
\begin{tabular}{ l c c c c c c}
\toprule
\multirow{2}{*}{\textbf{Model / Merge}}
& \textbf{Ratio} 
& \textbf{Math} 
& \textbf{Code} 
& \multicolumn{2}{c}{\textbf{General}} 
& \textbf{Instr./Agent} \\
\cmidrule(lr){3-3}
\cmidrule(lr){4-4}
\cmidrule(lr){5-6}
\cmidrule(lr){7-7}
& (src1:src2) 
& AIME25 
& LiveCode v5 
& MMLU-Pro* 
& GPQA-D 
& IFBench / $\tau^2$ \\ 

\hline
\rowcolor{gray!10}\multicolumn{7}{l}{{\textbf{\textit{(A) SFT--SFT Model Merging}}}} \\
\hline

(src1) SFT model 1 
& -- 
& \underline{66.00} 
& 43.91 
& \underline{78.29} 
& \underline{63.64} 
& 29.59 \\

(src2) SFT model 2 
& -- 
& 62.67 
& \underline{53.58} 
& 75.00 
& 50.51 
& \underline{42.52} \\ \hdashline

\multirow{3}{*}{Merged} 
& 8:2 
& 66.67 
& 46.42 
& 79.29 
& 64.14 
& 39.46 \\
 
& 5:5 
& \textbf{72.67} 
& 51.61 
& \textbf{80.43} 
& \textbf{69.19} 
& 50.00 \\

& 2:8 
& 66.00 
& \textbf{53.76} 
& 78.71 
& 59.09 
& \textbf{53.40} \\

\hline
\rowcolor{gray!10}\multicolumn{7}{l}{\textit{\textbf{(B) General--Agent Model Merging}}} \\
\hline
(src1) General Model 
& -- 
& \underline{70.67} 
& \underline{52.69} 
& \underline{78.00} 
& \underline{71.21} 
& 52.38 / 21.90 \\

(src2) Agentic Task Model 
& -- 
& 40.00 
& 10.22 
& 75.29 
& 57.07 
& 38.78 / \underline{65.80} \\\hdashline

\multirow{3}{*}{Merged} 
& 8:2 
& 70.00 
& 50.54 
& 76.86 
& \textbf{75.25} 
& \textbf{53.06} / 66.67 \\

& 5:5 
& 60.00 
& 45.34 
& \textbf{79.14} 
& 70.71 
& 51.70 / \textbf{76.32} \\

& 2:8 
& 45.33 
& 24.55 
& 77.00 
& 65.15 
& 43.54 / 73.10 \\
\bottomrule
\end{tabular}
}
\caption{Performance across different model merging ratios. 
(A) Merging two SFT models. 
(B) Merging a general-purpose model with an agentic model, with emphasis on agentic-task performance. 
An asterisk (*) indicates evaluation on a limited subset of the benchmark.}
\label{tab:merged_model_merging}
\end{table*}

Model merging offers an efficient post-training mechanism for improving performance without incurring additional optimization cost~\cite{li2025model, tian2025wsm, rame2024warp}. Rather than retraining a single model to simultaneously satisfy heterogeneous objectives, merging enables capability integration directly in parameter space, making it particularly suitable for large-scale LLM post-training pipelines~\cite{cohere2025command, bak2025kanana}.


In Mi:dm K 2.5 Pro, we first train multiple domain-specialized models during SFT, each emphasizing different capabilities such as mathematics, code, and general knowledge through distinct data mixtures and training ratios. We then merge these models to consolidate complementary strengths. As shown in \cref{tab:merged_model_merging}(A), SFT merging reduces task-wise variance and produces a more uniformly balanced model across benchmarks.

To further strengthen agentic capabilities, we additionally train the model specialized for agent behaviors and tool-use scenarios and merge them into the tool-use SFT model. As illustrated in \cref{tab:merged_model_merging}(B), this approach substantially improves agent-task performance while constraining degradation in mathematics and code. The merging coefficients further provide an explicit and controllable trade-off between agentic gains and general-purpose retention.

We adopt linear merging as a simple yet effective parameter-space aggregation strategy for models derived from a shared pre-trained backbone. This enables us to unify diverse domain capabilities into a single model without additional training overhead. The merged model then serves as a stable foundation for subsequent reinforcement learning, which further improves reasoning consistency and overall response reliability.
\paragraph{Reasoning RL Training Strategy.}
Following SFT-based model merging, the RL stage aims to mitigate output instability that may occur during the merging process and to refine the consistency and coherence of reasoning traces. To achieve this, we construct the training data centered around the RLVR based training methodology~\cite{guo2025deepseek, liu2024deepseek} and further optimize the model using these signals. We conduct RL training under an on-policy reinforcement learning configuration~\cite{schulman2017proximal}. At each update step, we sample 16 responses per prompt for 128 prompts and use the resulting trajectories as training signals. This configuration helps stabilize the output distribution of the model after merging while enabling relative quality comparisons across various response traces.


This stage extends instruction following beyond mathematics, code, and STEM by incorporating data designed to strengthen the execution of complex instruction. It additionally includes tasks that require structurally correct outputs, such as agentic tool use and JSON schema generation. For tool-use optimization, we adopt a two-track strategy. In RLVR, we focus on preventing regression in structured tool-call generation to maintain execution correctness. In parallel, we apply Direct Preference Optimization (DPO) to improve tool selection accuracy from a predefined tool inventory. We then merge the two specialized variants to combine structural robustness with precise selection capability.

\cref{tab:reasoning_data_mix} summarizes the domain-specific mixture ratios. As in the SFT stage~\cref{subsubsec: Reasoning SFT}, we apply batch-level blending to mix prompts from each domain within each training batch according to predefined ratios.

In the reinforcement learning stage, we jointly optimize the training framework and system configuration to stabilize post-merge generation while improving efficiency for long-horizon reasoning. We use \texttt{verl}~\cite{sheng2024hybridflow} as the RL framework and adopt a GSPO-based fully asynchronous training strategy~\cite{fu2025areal, han2025asyncflow} that decouples training (trainer) from generation (rollouter). This architecture reduces system bottlenecks caused by large variance in response lengths, and \cref{subsec:rl_efficiency_curriculum} details the efficiency and stabilization techniques used.

Additionally, we apply a shared penalty scheme to improve the efficiency and stability of reward computation. For every prompt, we assign a format penalty and a repetition penalty. When a reasoning trace fails to terminate properly, when special tokens repeat excessively and prevent the output from being parsed, or when specific words or phrases are repeated excessively, we skip accuracy-based reward computation and instead assign a negative reward. These penalties provide a strong constraint signal against failure modes that must be avoided, reducing reliance on computationally expensive accuracy-based rewards and improving both training stability and overall efficiency.

We compute task specific rewards solely based on the accuracy of the final response, independent of the common penalty and reasoning budget. For code tasks, we prompt the code blocks to be wrapped with \verb+```+ through system prompts and assign continuous rewards between 0 and 1 based on the pass rate of predefined test cases for Python code extracted according to that pattern. For STEM tasks, we evaluate problems by considering their formal characteristics and calculate the rewards based on whether parsed results match the correct answer according to the specified output format for multiple choice responses. For IF tasks, we use the evaluation code from IFBench~\cite{pyatkin2025generalizing} to measure instruction compliance at each instruction unit as rewards. In structured output tasks, we compute rewards by validating the structural consistency of outputs against JSON schemas and by providing type definitions as input. We normalize all accuracy based rewards to the $[-1,1]$ range to prevent specific sequences from having excessive advantage.

We employ GSPO~\cite{zheng2025groupsequencepolicyoptimization} as the RL algorithm. Unlike GRPO, GSPO optimizes policies at the sequence level, which reduces token-level variance and mitigates log-probability instability during long-sequence generation. This improves overall training stability.

\noindent ~\cref{tab:reasoning-rl-stagewise_performance} presents the performance change results in the early stages of reasoning post-training.

\begin{table*}[t]
\centering
\setlength{\tabcolsep}{5pt}
\small
\renewcommand{\arraystretch}{1.7}
\resizebox{0.85\textwidth}{!}{%
\begin{tabular}{l c c c c c c c c}
\toprule
\multirow{2}{*}{\textbf{Model}}
& \multicolumn{3}{c}{\textbf{Math}}
& \multicolumn{2}{c}{\textbf{Coding}}
& \multicolumn{3}{c}{\textbf{General}} \\
\cmidrule(lr){2-4}
\cmidrule(lr){5-6}
\cmidrule(lr){7-9}
& AIME25 & MATH500 & Math-H
& LCB & \makecell{Human\\Eval+}
& \makecell{MMLU\\-Pro} & \makecell{KMMLU\\-Redux} & GPQA-D \\
\midrule
\makecell[l]{Reasoning SFT\\(Stage 1)}
& 65.33 & 91.20 & 90.56
& 43.01 & 89.02
& 67.14 & -- & 61.11 \\ \hdashline
\makecell[l]{Reasoning SFT\\(+Merge)}
& 65.33 & {91.80} & {93.35}
& 51.79 & 89.63
& {79.43} & {65.67} & 67.17 \\\hdashline
Reasoning RL
& {70.67} & 90.60 & 92.37
& {52.69} & {90.85}
& 78.00 & 65.56 & {71.21} \\
\bottomrule
\end{tabular}%
}
\caption{Stage-wise performance changes during reasoning-focused post-training}
\label{tab:reasoning-rl-stagewise_performance}
\end{table*}
\paragraph{Fusion Training Strategy.}
\begin{figure}[h!]
  \centering
  \includegraphics[width=0.9\linewidth]{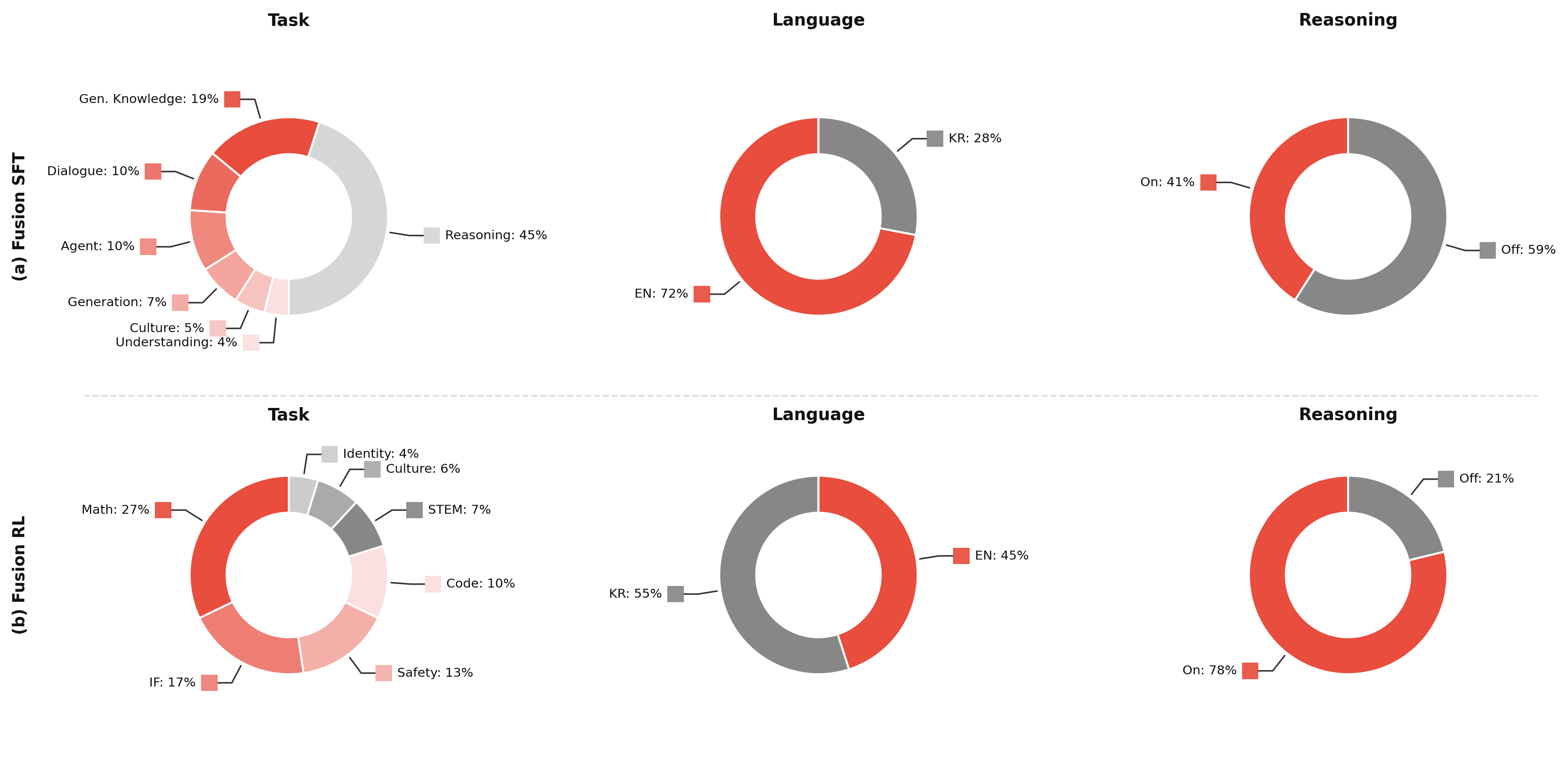}
  \caption{Training data composition of Fusion SFT and Fusion RL across task, language, and reasoning dimensions.}
   \label{fig:fusion_sft_rl_data_dist}
\end{figure}

\begin{table*}[h!]
\centering
\renewcommand{\arraystretch}{1.2}
\setlength{\tabcolsep}{5pt}

\resizebox{\textwidth}{!}{%
\begin{tabular}{lc *{9}{c}}
\toprule
\multirow{3}{*}{\textbf{Model}}
& \multirow{3}{*}{\textbf{Reasoning}}
& \multicolumn{6}{c}{\textbf{English}}
& \multicolumn{3}{c}{\textbf{Code}} \\
\cmidrule(lr){3-8}\cmidrule(lr){9-11}
& & \multicolumn{2}{c}{\textbf{Instr. Following}}
  & \textbf{Gen. Rsn.}
  & \multicolumn{2}{c}{\textbf{Math}}
  & \textbf{Gen. Know.}
  & \multirow{3}{*}{\makecell{LCB\\V6\\(pass@1)}}
  & \multirow{3}{*}{\makecell{Human\\Eval+\\(pass@1)}}
  & \multirow{3}{*}{\makecell{MBPP+\\(pass@1)}} \\
\cmidrule(lr){3-4}\cmidrule(lr){5-5}\cmidrule(lr){6-7}\cmidrule(lr){8-8}
& & \makecell{IFEval\\(avg)}
  & \makecell{IFBench\\(EM)}
  & \makecell{GPQA-D\\(EM)}
  & \makecell{MATH-H\\(EM)}
  & \makecell{AIME25\\(EM)}
  & \makecell{MMLU-Pro\\(EM)} \\
\hline
\rowcolor{gray!10}\multicolumn{11}{l}{\textit{\textbf{(A) English \& Code}}} \\
\hline
\multirow{2}{*}{\quad Reasoning Model}
  & Off & 85.04 & --    & --    & 74.24 & --    & 70.57 & --    & 83.54 & 87.04 \\
  & On  & 88.43 & 48.64 & 67.68 & 95.24 & 61.33 & 79.14 & 46.26 & 89.02 & 89.42 \\
\midrule
\multirow{2}{*}{\quad Fusion SFT}
  & Off & 88.26 & --    & --    & 50.68 & --    & 73.43 & --    & 75.00 & 81.75 \\
  & On  & 81.44 & 53.06 & 56.06 & 90.41 & 44.67 & 74.71 & 16.74 & 85.98 & 82.80 \\
\midrule
\multirow{2}{*}{\quad Fusion SFT Merge}
  & Off & 85.10 & --    & --    & 73.34 & --    & 69.86 & --    & 80.49 & 90.21 \\
  & On  & 87.94 & 49.66 & 67.68 & 95.24 & 66.00 & 78.86 & 47.36 & 90.85 & 89.95 \\
\midrule
\multirow{2}{*}{\quad Fusion RL}
  & Off & 85.59 & --    & --    & 79.76 & --    & 71.14 & --    & 78.66 & 87.83 \\
  & On  & 89.56 & 56.46 & 64.14 & 95.17 & 68.00 & 78.43 & 50.22 & 91.46 & 89.68 \\
\bottomrule
\end{tabular}%
}
\resizebox{\textwidth}{!}{%
\begin{tabular}{lc *{9}{c}}
\hline
\multirow{3}{*}{\textbf{Model}}
& \multirow{3}{*}{\textbf{Reasoning}} & \textbf{Instr. Following}
  & \multicolumn{4}{c}{\textbf{General Knowledge}}
  & \multicolumn{2}{c}{\textbf{Math}}
  & \multicolumn{2}{c}{} \\
\cmidrule(lr){3-3}\cmidrule(lr){4-7}\cmidrule(lr){8-9}
& & \makecell{{Ko-IFEval}\\(avg)}
  & \makecell{{KoBALT}\\(EM)}
  & \makecell{{CLIcK-L}\\(EM)}
  & \makecell{{CLIcK-C}\\(EM)}
  & \makecell{{KMMLU}\\(EM)}
  & \makecell{{AIME25-Ko}\\(EM)}
  & \makecell{{HRM8K}\\(EM)}
  & \multicolumn{2}{c}{} \\
\hline
\rowcolor{gray!10}\multicolumn{11}{l}{\textit{\textbf{(B) Korean}}} \\
\hline
\multirow{2}{*}{\quad Reasoning Model}
  & Off & 80.59 & 37.43 & 73.85 & 78.29 & --    & --    & --    & & \\
  & On  & --    & --    & --    & --    & 71.24 & 64.67 & 80.74 & & \\
\midrule
\multirow{2}{*}{\quad Fusion SFT}
  & Off & 86.10 & 35.57 & 70.92 & 90.48 & 66.00 & 78.14 & --    & & \\
  & On  & --    & --    & --    & --    & --    & --    & 64.36 & & \\
\midrule
\multirow{2}{*}{\quad Fusion SFT Merge}
  & Off & 82.24 & 38.57 & 71.08 & 79.11 & --    & --    & --    & & \\
  & On  & --    & --    & --    & --    & 71.74 & 71.33 & 81.35 & & \\
\midrule
\multirow{2}{*}{\quad Fusion RL}
  & Off & 82.80 & 44.71 & 73.54 & 81.71 & --    & --    & --    & & \\
  & On  & --    & --    & --    & --    & 71.86 & 71.33 & 81.77 & & \\
\bottomrule
\end{tabular}
}

\caption{Performance comparison across fusion training stages on English, Code, and Korean benchmarks.}
\label{tab:overall_benchmarks}
\end{table*}

Building upon the robust reasoning and task-solving foundations established in previous stages, we conduct Fusion Training—comprising Fusion SFT and Fusion RL—to refine the model's practical usability. While prior iterations focus predominantly on reasoning modes, this stage ensures that the model yields appropriate responses in non-reasoning contexts as well.


In the Fusion SFT phase, we partition the training data into reasoning and non-reasoning subsets. We deliberately emphasize non-reasoning tasks that are relatively underrepresented in earlier stages, including creative writing, translation, and general question answering. As shown in ~\cref{fig:fusion_sft_rl_data_dist}, we raise the ratio of non-reasoning subsets in the data mixture to favor instilling versatile conversational alignment. To ensure training stability, we also adopt a balanced batch design that preserves the predefined reasoning-to–non-reasoning ratio within each mini-batch, consistent with the strategy used in the reasoning stage.

To mitigate the performance degradation observed on certain benchmarks during the initial SFT phase, we reintroduce the model merging strategy, which has demonstrated effectiveness in the reasoning SFT stage. This approach enables us to identify an appropriate trade-off between raw reasoning capability and conversational fluency prior to reinforcement learning.

The Fusion RL stage aims to improve controllability, reinforce model identity, and mitigate tool-related hallucinations. We design a more fine-grained reward system that promotes high-fidelity and deployment-ready outputs.

First, we introduce strict penalties for mode–channel mismatches (e.g., invoking an analysis channel for a non-reasoning task) and bypass accuracy evaluation for incorrectly formatted responses, ensuring that structural violations are penalized before performance assessment and thereby enforcing structural consistency.

\quad Second, to enhance reliability in real-world settings, we incorporate penalties targeting unintended language switching (code-mixing) and tool-use hallucinations. These constraints reduce spurious tool invocation and unauthorized language shifts, improving robustness under real-world usage conditions.

Finally, we replace the conventional reward model-based RLHF pipeline with an LLM-as-a-judge framework. Rewards are computed asynchronously and overlapped with response generation, minimizing additional latency while maintaining training efficiency described in~\cref{subsubsec:LLM-based Reward Model Development.}.

The results in~\cref{tab:overall_benchmarks} indicate that Fusion Training progressively rebalances reasoning and non-reasoning capabilities without compromising core reasoning strength. Fusion SFT improves conversational alignment, model merging stabilizes performance regressions observed during initial rebalancing, and Fusion RL further enhances controllability and robustness under practical usage conditions.

\subsection{RL Training Efficiency and Operational Design}
\label{subsec:rl_efficiency_curriculum}
In this section, we summarize the design choices that improve learning efficiency and stability during the reinforcement learning stage. Rather than modifying the underlying RL algorithm, we focus on operational components, including difficulty-aware data selection and curriculum learning, system-level efficiency enabled by an asynchronous training architecture, and LLM-assisted reward-signal generation.

\paragraph{Difficulty-aware Prompt Selection.}      
RL is a key technique for aligning language models with human preferences and values, and in particular, on-policy reinforcement learning uses only data generated during the training process, so learning efficiency directly correlates to performance. However, not all prompts provide the same learning signal, and inefficient data usage can hinder performance improvement under limited resources and time. Therefore, we outline three key challenges to improve the efficiency of on-policy reinforcement learning. First, repeated learning on easy prompts that the model has already mastered, or the use of prompts that are excessively difficult at the current level, can reduce learning efficiency, therefore making data selection at a difficulty level appropriate for the model becomes necessary~\cite{bengio2009curriculum,soviany2022curriculum}. Second, multiple objectives such as safety, helpfulness, and format compliance can conflict with each other~\cite{bai2022training,askell2021general}, and the method used to integrate these reward signals directly affects alignment quality. Third, while traditional reward models incur high training and maintenance costs and remain vulnerable to changes in evaluation criteria, the LLM-as-a-Judge approach offers flexibility but also introduces practical challenges related to cost and evaluation consistency~\cite{lee2023rlaif,liu2023g}.

To mitigate these issues, we adopt a curriculum learning strategy. We define prompt difficulty relative to the current policy—i.e., how challenging it is for the policy to produce a high-quality response—and treat it as a dynamic quantity that evolves over training. To address mismatched difficulty criteria across data sources, we re-define task and difficulty labels under a unified internal rubric and adjust the difficulty distribution to match the training-stage characteristics of the model. In STEM in particular, RLVR with verifiable rewards has been shown to be effective; accordingly, we apply difficulty-aware prompt selection more aggressively.

In the mathematics domain, we define five difficulty levels (Levels 1–5) and categorize problems into seven content-based subdomains. After rebalancing the data mixture using difficulty and subdomain tags, we maintain or improve performance while using only 20.1\% of the full dataset. Notably, MATH500 improves by more than 3×, and the high-difficulty benchmarks AIME 2024 and AIME 2025 show additional gains of +20.0\% and +28.6\%, respectively.
In the code domain, we first obtain a coarse difficulty estimate using simple heuristics (number of test cases and problem-statement length), and then assign a 1.0–5.0 difficulty score via an LLM-based assessment that jointly considers problem comprehension difficulty, reasoning depth, implementation burden, and related factors. \cref{tab:difficulty_factors} lists the assessment criteria for code difficulty, and \cref{tab:difficulty_score_bins} summarizes the distribution across score bins. Overall, this difficulty-aware curriculum reduces unnecessary training while improving both the efficiency and stability of reinforcement learning under constrained resources.

\begin{table}[h!]
\centering
\renewcommand{\arraystretch}{1.15}
\resizebox{\textwidth}{!}{
\begin{tabular}{p{5.5cm} p{7.5cm} p{3.5cm}}
\toprule\midrule\rowcolor{gray!10}
\textbf{Assessment Factor} & \textbf{Description} & \textbf{Difficulty Contribution} \\
\midrule
Algorithmic Complexity
& Required algorithm types: Brute-force, Greedy, DP, Graph, advanced data structures, etc.
& when advanced algorithms is required $\uparrow$ \\ \hdashline

Reasoning Depth
& Number of logical and computational steps required for problem solving    
& when multi-step reasoning is needed  $\uparrow$ \\\hdashline

Edge case Richness
& Existence of ambiguous or non trivial corner cases     
& when complex edge cases are abundant $\uparrow$ \\\hdashline

Constraint Pressure
& Optimization necessary according to input size      
& when input size is large $\uparrow$ \\\hdashline

Implementation Load
& Degree to which meticulous coding or mathematical approach is required        
& when complex implementation is needed $\uparrow$ \\\hdashline

Conceptual Abstraction
& Whether intuitive logic vs.\ advanced concepts (mathematical theorems, specialized techniques, etc.) are required
& when advanced concepts are required $\uparrow$ \\
\bottomrule
\end{tabular}
}
\caption{Code domain problem difficulty components and contributing factors}        
\label{tab:difficulty_factors}
\end{table}

\begin{table}[h!]
\centering
\renewcommand{\arraystretch}{1.15}
\resizebox{\textwidth}{!}{
\begin{tabular}{p{3.5cm} p{4.5cm} p{8.5cm}}
\toprule\midrule\rowcolor{gray!10}
\textbf{Score} & \textbf{Difficulty} & \textbf{Characteristics} \\
\midrule
1.0 -- 1.5
& Very Easy
& Simple implementation, basic syntax level \\\hdashline

1.5 -- 2.5
& Easy
& Basic algorithms (sorting, search), simple condition handling \\\hdashline

2.5 -- 3.5
& Medium
& Intermediate algorithms (basic DP, BFS/DFS), multiple condition handling \\\hdashline

3.5 -- 4.5
& Difficult
& Advanced algorithms (complex DP, graph optimization), complex reasoning \\\hdashline

4.5 -- 5.0
& Very Difficult
& Highly advanced algorithms, mathematical insight, combination of optimization techniques required \\
\bottomrule
\end{tabular}
}
\caption{Code domain difficulty definition by score range}      
\label{tab:difficulty_score_bins}
\end{table}

\paragraph{Asynchronous RL Execution and Stabilization.} 
In reinforcement learning processes involving long form reasoning tasks, not only training stability but also efficient utilization of system resources directly impacts overall training performance. Particularly in environments with high variance in response length, the execution mode of the training framework can introduce significant inefficiencies.

In conventional synchronous RL, the system advances to the next update only after rollouts have been generated for all prompts in the batch. For reasoning workloads with high variance in response length, this induces substantial straggler wait: the longest generation dominates step time, leaving many GPUs idle.
By contrast, a fully asynchronous RL architecture decouples rollout generation from training and runs them in parallel. A rollouter continuously generates trajectories under a (slightly stale) policy and writes them to a buffer, while a trainer asynchronously consumes buffered trajectories and performs parameter updates. The rollouter is periodically refreshed to track the latest policy, enabling sustained parallelism between generation and learning. This design mitigates the long-tail straggler bottleneck common in synchronous RL and significantly reduces GPU idle time.

To quantitatively analyze the system-efficiency impact of execution mode for GSPO training, we compare synchronous RL and fully asynchronous RL under the same model configuration on an 8-node setup.
\cref{tab:training_mode_efficiency} presents the step time breakdown and token throughput for both training modes. In the fully asynchronous configuration, the system reduces the time required for the rollout generation stage to less than half compared to the synchronous method, which demonstrates that the system effectively alleviates the long tail bottleneck caused by varying response length. Although the actor update time increases, it has limited impact on the overall step time since the system performs it overlapping with the rollout generation. As a result, fully asynchronous GSPO training reduces step time by approximately 15\% and improves token throughput by approximately 30\% compared to the synchronous baseline. These improvement effects become more pronounced as the maximum response length increases and as the scale of resources used for training grows larger.

\begin{table}[h!]
\centering
\resizebox{\textwidth}{!}{
\begin{tabular}{l c c c c c}
\toprule
\textbf{Training Mode} 
& \textbf{Resource Allocation} 
& \textbf{Step (sec)} 
& \textbf{Generation (sec)} 
& \textbf{Update Actor (sec)} 
& \textbf{Throughput} \\
\midrule
Synchronous 
& 64 
& 388.72 
& 286.83 
& 50.91 
& 211.31 \\
Fully asynchronous 
& 32 / 32 
& 332.61 
& 124.08 
& 119.35 
& {275.37} \\
\bottomrule
\end{tabular}
}
\caption{Comparison of Colocate sync and Fully async training speed}        
\label{tab:training_mode_efficiency}
\end{table}
However, in asynchronous training, trajectories generated by the rollouter may become off-policy relative to the trainer’s latest policy. In long-form reasoning tasks, such policy lag can lead to training instability or even collapse. To mitigate this, we apply rollout correction~\cite{yao2025on, espeholt2018impala}. Specifically, for each sequence generated by the rollouter, we compare its log probability under the rollouter policy at generation time with its log probability under the current training policy; if the discrepancy exceeds a predefined threshold, we drop the sequence from training. This filtering limits off-policy–induced distribution shift and improves training stability.
\paragraph{LLM-based Reward Model Development.} 
\label{subsubsec:LLM-based Reward Model Development.}
The approach of constructing reward models for individual attributes such as safety and helpfulness has the advantage that the trained model can internalize the preferences of human evaluators and provide relatively consistent reward signals. However, this approach entails several practical limitations. Training reward models requires large scale preference data collection and a separate training process, and the system inevitably requires retraining when new evaluation criteria are added or existing criteria are modified. Additionally, Moreover, reward models are largely opaque, limiting interpretability of their judgments. To address these limitations, we adopt the LLM-as-a-Judge paradigm and use LLMs directly to generate reward signals.

To use LLMs as reward models, it is essential to define evaluation criteria clearly and to design evaluation prompts that elicit consistent and calibrated judgments~\cite{wang2023self}. We begin by explicitly framing the LLM as an expert evaluator, encouraging assessments from a stable perspective and leveraging domain-specific judgment. We further improve consistency by moving beyond abstract notions and enumerating concrete sub-criteria. For example, helpfulness can be operationalized in terms of accuracy, completeness, relevance, and clarity.

We also include a reasoning protocol that encourages the LLM to follow a stepwise deliberation process before issuing a final judgment. By explicitly generating evaluation rationales, this improves the stability and reproducibility of the judgments. To quantify outcomes, we provide a scoring rubric that defines the rating scale and the criteria for each score level. Depending on task characteristics, we choose either absolute or comparative scoring and adjust the score range and granularity accordingly. Finally, we incorporate a self-verification step in which the LLM revisits its initial assessment, mitigating evaluation errors and further improving judgment quality.


Additionally, to enable reliable use of LLM evaluation outputs in downstream pipelines, we adopt structured output formats~\cite{openai24structured}. Each structured output includes a score field representing the final reward value and a reasoning field that records the evaluation rationale; for multi-dimensional assessments, we also include per-dimension sub-scores. Because safety evaluation has clear decision criteria, we use discrete judgments, whereas helpfulness is scored on a continuous scale to capture finer-grained quality differences. The reasoning field not only improves interpretability of reward signals but also supports debugging and quality auditing.

Under these configurations, we compare human-tagged evaluations with LLM-based reward signals and find an agreement rate of 91\%. This indicates that the proposed reward-signal generation method is sufficiently reliable for practical use.

\section{Evaluation}
\label{section5:evalution}

\subsection{Quantitative Evaluation}
\label{subsec:quantitative_evaluation}

We conduct a quantitative evaluation using a benchmark set that combines representative public benchmarks, translated versions of those public benchmarks in Korean, and in-house proprietary benchmarks constructed for Korean language and domain specific evaluation. 
We compare our model with both globally competitive models and industry leading Korean-specialized models; an overview is provided in \cref{tab:model_overview_baselines}.

The evaluation is organized along two complementary axes: English benchmarks for general-purpose capabilities and Korean benchmarks for Korean-specific comprehension, culturally grounded reasoning, and knowledge of Korean society and context. Together, these evaluations provide a broad view of the model's performance across both general and Korean-specific settings.

In addition, we examine statistical significance for selected comparisons to support more reliable interpretation of performance differences.
\begin{table}[h!]
\centering
\renewcommand{\arraystretch}{1.15}
\small
\setlength{\tabcolsep}{6pt}
\resizebox{0.8\textwidth}{!}{%
\begin{tabular}{lccccc}
\toprule
& \multirow{2}{*}{\textbf{Mi:dm K Pro}}
& \textbf{HyperCLOVAX} 
& \textbf{Qwen3-30B-A3B} 
& \textbf{Solar-Open} 
& \textbf{K-EXAONE} \\
&  
& \textbf{SEED-Think-32B} 
& \textbf{Thinking} 
& \textbf{100B} 
& \textbf{236B-A23B} \\
\midrule
\textbf{Developer} 
& KT
& NAVER Cloud 
& Alibaba Qwen 
& Upstage 
& LG AI Research \\
\textbf{Architecture} 
& Dense
& Dense
& MoE 
& MoE 
& MoE \\
\textbf{\# Total Params} 
& 32B 
& 32B 
& 30.5B 
& 102B 
& 236B \\
\textbf{\# Activated Params} 
& 32B 
& 32B 
& 3.3B 
& 12B 
& 23B \\
\bottomrule
\end{tabular}%
}
\caption{Overview of open or externally reported baseline models used in our experiments.}
\label{tab:model_overview_baselines}
\end{table}

\paragraph{General English Benchmark.}
To evaluate the general capabilities of Mi:dm K 2.5 Pro, we use a representative set of public English benchmarks covering reasoning, knowledge, mathematics, instruction following, coding, and agentic capability. Selected results are additionally aligned with externally reported evaluations from Artificial Analysis~\cite{aaii_2025}, providing an externally comparable view of performance on widely used public benchmark settings.

\begin{table*}[h!]
\centering
\renewcommand{\arraystretch}{1.2}
\setlength{\tabcolsep}{5pt}
\resizebox{0.85\textwidth}{!}{%
\begin{tabular}{lc *{6}{c}}
\toprule
\multirow{3}{*}{\textbf{Model}}
& \multirow{3}{*}{\textbf{Reasoning}}
& \multicolumn{6}{c}{\textbf{English}} \\
\cmidrule(lr){3-8}
& 
& \textbf{Gen. Rsn.}
& \textbf{Gen. Know.}
& \multicolumn{2}{c}{\textbf{Math}}
& \multicolumn{2}{c}{\textbf{Instr. Following}} \\
\cmidrule(lr){3-4}\cmidrule(lr){5-6}\cmidrule(lr){7-8}
& 
& \makecell{GPQA-D*\\(EM)}
& \makecell{MMLU-Pro\\(EM)}
& \makecell{MATH-H\\(EM)}
& \makecell{AIME25\\(EM)}
& \makecell{IFEval\\(avg)}
& \makecell{IFBench*\\(EM)} \\
\hline
\rowcolor{gray!10}\multicolumn{8}{l}{\textit{\textbf{(A) English}}} \\
\hline
\multirow{2}{*}{\quad K-EXAONE-236B-A23B}
& Off  & 70 & 80.96 & 91.00 & 40    & 88.40 & 40 \\ 
& On   & 78 & 83.40 & 97.36 & 86.67 & 92.90 & 65 \\
\hdashline
\multirow{2}{*}{\quad Solar-Open-100B}
& Off  & -- & --    & --    & --    & --    & -- \\
& On   & 66 & 79.77 & 95.92 & 76.67 & 89.42 & 58 \\
\midrule
\multirow{2}{*}{\quad Qwen-3-30B-A3B}
& Off  & 66 & 78.67 & 92.70 & 63.30 & 89.00 & 33 \\
& On   & \underline{71} & \underline{80.10} & \textbf{97.05} & \textbf{86.67} & \textbf{91.50} & \textbf{51} \\
\hdashline
\multirow{2}{*}{\quad HyperCLOVAX-SEED-Think-32B}
& Off  & -- & 67.31 & 62.50 & 3.30  & 86.83 & -- \\
& On   & 62 & 78.50 & 95.17 & 56.67 & 86.40 & 38 \\
\hdashline
\multirow{2}{*}{\quad Mi:dm K 2.5 Pro (March `26)}
& Off  & -- & 73.98 & 72.20 & 33.30 & 86.23 & -- \\
& On   & \textbf{72} & \textbf{81.80} & \underline{96.60} & \underline{70.00} & \underline{89.50} & \textbf{51} \\
\bottomrule
\end{tabular}%
}
\resizebox{0.85\textwidth}{!}{%
\begin{tabular}{lc *{5}{c}}
\toprule
\multirow{2}{*}{\textbf{Model}}
& \multirow{2}{*}{\textbf{Reasoning}}
& \multicolumn{3}{c}{\textbf{Code}}
& \multicolumn{2}{c}{\textbf{Agentic}} \\
\cmidrule(lr){3-5}\cmidrule(lr){6-7}
&
& \makecell{LCB\\V6\\(pass@1)}
& \makecell{Human\\Eval+\\(pass@1)}
& \makecell{MBPP+\\(pass@1)}
& \makecell{Terminal*\\Bench\\(pass@1)}
& \makecell{$\tau^2$-Bench*\\Telecom\\(pass@1)} \\
\hline
\rowcolor{gray!10}\multicolumn{7}{l}{\textit{\textbf{(B) Code \& Agentic}}} \\
\hline
\multirow{2}{*}{\quad K-EXAONE-236B-A23B}
& Off  & 57.10 & 83.50 & 66.90 & 7  & 59 \\
& On   & 89.00 & 90.85 & 97.88 & 23 & 74 \\
\hdashline
\multirow{2}{*}{\quad Solar-Open-100B}
& Off  & --    & --    & --    & -- & -- \\
& On   & 72.80 & 92.07 & 75.13 & 2  & 48 \\
\midrule
\multirow{2}{*}{\quad Qwen-3-30B-A3B}
& Off  & 58.30 & 88.40 & 77.70 & 7  & 22 \\
& On   & \textbf{78.96} & {87.20} & \textbf{90.48} & 5 & 28 \\
\hdashline
\multirow{2}{*}{\quad HyperCLOVAX-SEED-Think-32B}
& Off  & 28.10 & 74.30 & 89.90 & -- & -- \\
& On   & 68.34 & \underline{88.41} & 83.07 & \textbf{12} & \underline{87} \\
\hdashline
\multirow{2}{*}{\quad Mi:dm K 2.5 Pro (March `26)}
& Off  & 40.10 & 82.90 & 76.70 & -- & -- \\
& On   & \underline{74.79} &\textbf{92.07} & 89.68  & 3 & \textbf{89} \\
\bottomrule
\end{tabular}%
}
\caption{Performance comparison across English, code, and agentic benchmarks. * denotes results reported from Artificial Analysis~\cite{aaii_2025}.}
\label{tab:overall_benchmarks}
\end{table*}

The English benchmark set is organized into six capability categories:
\begin{itemize}
    \item \textbf{Reasoning} -- \textsc{GPQA-Diamond}~\cite{rein2024gpqa}
    \item \textbf{General Knowledge} -- \textsc{MMLU-Pro}~\cite{wang2024mmlu}
    \item \textbf{Mathematics} -- \textsc{MATH-Hard}~\cite{hendrycksmath2021} and \textsc{AIME25}~\cite{maa_aime_2025}
    \item \textbf{Instruction Following} -- \textsc{IFEval}~\cite{zhou2023instructionfollowingevaluationlargelanguage} and \textsc{IFBench}~\cite{pyatkin2025generalizing}
    \item \textbf{Coding} -- \textsc{LiveCodeBench v6}~\cite{jain2024livecodebench}, \textsc{HumanEval+}~\cite{evalplus}, and \textsc{MBPP+}~\cite{evalplus}
    \item \textbf{Agentic Capability} -- \textsc{Terminal-Bench}~\cite{merrill2026terminalbenchbenchmarkingagentshard} and \textsc{$\tau^2$-Bench Telecom}~\cite{barres2025tau}
\end{itemize}

As shown in \cref{tab:overall_benchmarks}, Mi:dm K 2.5 Pro shows a broadly comparable performance to similar sized models on the English benchmark set, with its clearest advantages appearing on reasoning- and knowledge-oriented evaluations. Under reasoning-enabled inference, it records the highest scores among similar sized models on GPQA-D and MMLU-Pro, while remaining competitive on MATH-H. These results indicate that Mi:dm performs favorably on benchmarks that require broad reasoning and knowledge application.

On code and agentic benchmarks, Mi:dm K 2.5 Pro shows competitive performance with particular strengths. Among similar sized models, it achieves the highest HumanEval+ score at 92.07\% and the highest $\tau^2$-Bench Telecom score at 89\%. These results suggest that Mi:dm does not uniformly lead across all code-oriented evaluations, but demonstrates clear strengths on practical code synthesis and telecom-oriented agentic tasks.

In particular, although Mi:dm K 2.5 Pro is substantially smaller than K-EXAONE-236B-A23B, its MATH-H and MMLU-Pro scores remain competitive with those of the much larger model. Its LiveCodeBench score also exceeds that of Solar-Open-100B, and its $\tau^2$-Bench Telecom score surpasses those of both Solar-Open-100B and K-EXAONE-236B-A23B. Taken together, these results indicate that Mi:dm K 2.5 Pro achieves strong performance on most benchmarks despite its smaller scale, while achieving performance levels comparable to much larger models on several key evaluations.

\paragraph{Korean Specific Benchmark.}
To evaluate the Korean-specific capabilities of Mi:dm K 2.5 Pro, we adopt a benchmark setting designed to capture linguistic, cultural, and contextual properties that are central to Korean. Existing evaluations often provide limited coverage of such properties, particularly for honorific usage, pragmatic interpretation, and culturally grounded reference resolution. To address these limitations, we combine publicly available Korean benchmarks with in-house benchmarks developed by KT. Detailed descriptions of the in-house benchmark design are provided in Mi:dm 2.0~\cite{shin2026mi}.

The Korean benchmark set is organized into six capability categories:
\begin{itemize}
    \item \textbf{Korean Comprehension} -- \textsc{Ko-Sovereign*} (language and literature), \textsc{K-Pragmatics\footnotemark}, \textsc{KoBALT}~\cite{shin2025kobalt}, and \textsc{CLIcK-L}~\cite{kim2024click}
    \item \textbf{Society \& Culture} -- \textsc{Ko-Sovereign\footnotemark} (culture, folklore, and society), \textsc{K-Referential\footnotemark}, and \textsc{CLIcK-C}~\cite{kim2024click}
    \item \textbf{Instruction Following} -- \textsc{Ko-IFEval}~\cite{zhou2023instructionfollowingevaluationlargelanguage}
    \item \textbf{Reasoning} -- \textsc{Ko-Winogrande}~\cite{kim2025open} and \textsc{HRMCR}~\cite{son2025multi}
    \item \textbf{Korean Knowledge} -- \textsc{KMMLU}~\cite{son-etal-2025-kmmlu} and \textsc{Ko-Sovereign\footnotemark}
    \item \textbf{Mathematics} -- Evaluated with \textsc{HRM8K}~\cite{ko2025understand} and \textsc{AIME25-Ko}~\cite{lighteval}
\end{itemize}
\footnotetext{KT proprietary benchmark, internally developed for Korean-specific evaluation}
\begin{table*}[h!]
\centering
\setlength{\tabcolsep}{4.2pt}
\renewcommand{\arraystretch}{1.12}
\small
\resizebox{0.9\textwidth}{!}{
\begin{tabular}{lcccccccc}
\toprule
\multirow{2}{*}{\textbf{Model}} 
& \multirow{2}{*}{\textbf{Reasoning}}
& \multicolumn{4}{c}{\textbf{Comprehension}}
& \multicolumn{3}{c}{\textbf{Society \& Culture}}\\
\cmidrule(lr){3-6}\cmidrule(lr){7-9}

& & \makecell{Ko-Sov.\\(EM)}& \makecell{K-Prag.\\(EM)} & \makecell{KoBALT\\(EM)} & \makecell{CLIcK-L\\(EM)}
& \makecell{Ko-Sov.\\(EM)} & \makecell{K-Ref.\\(EM)} & \makecell{CLIcK-C\\(EM)}\\
\hline
\rowcolor{gray!10}\multicolumn{9}{l}{\textit{\textbf{(A) Korean Comprehension and Society \& Culture}}} \\
\hline

\multirow{2}{*}{\quad K-EXAONE-236B-A23B}
& Off & 62.50 & 91.50 & 51.10 & 79.10 & 64.00 & 84.00 & 79.90  \\
& On  & 68.00 & 93.73 & 61.86 & 87.23 & 81.40 & 92.80 & 81.49  \\
\hdashline

\multirow{2}{*}{\quad Solar-open-100B}
& Off & -- & -- & -- & -- & -- & -- & -- \\
& On  & 63.50 & 93.53 & 48.86 & 81.38 & 64.90 & 90.00 & 77.70  \\
\midrule

\multirow{2}{*}{\quad Qwen-3-30B-A3B}
& Off & 58.00 & 88.70 & 35.40 & 72.90 & 49.30 & 82.40 & 66.60  \\
& On  & 59.50 & 89.80 & 40.14 & 78.15 & 70.90 & 86.00 & 68.40 \\
\hdashline
\multirow{2}{*}{\quad HyperCLOVAX-SEED-Think-32B}
& Off & 61.50 & 87.50 & 33.10 & 72.30 & 60.00 & 90.00 & 78.70  \\
& On  & \underline{68.00} & \textbf{92.47} & \underline{49.86} & \underline{80.15} & \underline{80.80} & \textbf{94.80} & \textbf{83.20}\\
\hdashline

\multirow{2}{*}{\quad Mi:dm K 2.5 Pro (March `26)}
& Off & 69.00 & 89.00 & 46.50 & 79.10 & 63.60 & 87.60 & 81.30 \\
& On  & \textbf{73.50} & \underline{91.60} & \textbf{57.86} & \textbf{84.92} & \textbf{82.20} & \underline{92.80} & \underline{82.83}  \\
\midrule
\hline

\rowcolor{gray!10}\multicolumn{9}{l}{\textit{\textbf{(B) Instruction Following, Reasoning, Knowledge, and Math}}} \\
\hline

\multirow{3}{*}{\textbf{Model}} 
& \multirow{3}{*}{\textbf{Reasoning}}
& \multicolumn{2}{c}{\textbf{Reasoning}}
& \multicolumn{2}{c}{\textbf{Knowledge}}
& \multicolumn{2}{c}{\textbf{Math}}
& \textbf{Instr. Following}
\\
\cmidrule(lr){3-4}\cmidrule(lr){5-6}\cmidrule(lr){7-8}\cmidrule(lr){9-9}
& & \makecell{Ko-Winogrande\\(EM)} & \makecell{HRMCR\\(EM)}
& \makecell{KMMLU\\(EM)} & \makecell{Ko-Sov.\\(EM)}
& \makecell{HRM8K\\(EM)} & \makecell{AIME25-Ko\\(EM)} &\makecell{Ko-IFEval\\(avg)} \\
\midrule
\multirow{2}{*}{\quad K-EXAONE-236B-A23B}
& Off  & 81.80 & 9.00  & 74.91 & 69.00 & 80.40 & 43.30 & 87.45  \\
& On   & 86.02 & 36.00 & 77.90 & 71.90 & 89.08 & 86.67 & 91.00  \\
\hdashline

\multirow{2}{*}{\quad Solar-open-100B}
& Off  & -- & -- & -- & -- & -- & -- & --  \\
& On   & 84.13 & 40.00 & 72.58 & 65.80 & 81.11 & 76.67 & 88.19 \\
\midrule
\multirow{2}{*}{\quad Qwen-3-30B-A3B}
& Off  & 80.60 & 4.00  & 63.54 & 59.10 & 74.10 & 50.00 & 85.13 \\
& On   & \underline{84.13} & \textbf{35.00} & 72.40 & 61.80 & 87.49 & \textbf{86.67 }& \textbf{93.20}  \\
\hdashline

\multirow{2}{*}{\quad HyperCLOVAX-SEED-Think-32B}
& Off  & 75.40 & 9.00  & 65.53 & 63.00 & 56.00 & 3.30  & 79.74 \\
& On   & 81.68 & 22.00 & \underline{74.80} & \underline{70.70} & 79.53 & 40.00 & 84.20  \\
\hdashline
\multirow{2}{*}{\quad Mi:dm K 2.5 Pro (March `26)}
& Off  & 79.40 & 16.00 & 71.06 & 69.60 & 64.30 & 16.60 & 81.03 \\
& On   & \textbf{86.20} & \textbf{35.00} & \textbf{76.90} & \textbf{71.60} & \underline{86.83} & \underline{70.00} & \underline{85.60}  \\
\bottomrule
\end{tabular}%
}
\caption{Performance comparison across Korean benchmark groups. The upper block reports comprehension, society \& culture, and generation benchmarks, while the lower block reports instruction following, reasoning, knowledge, and math benchmarks.}
\label{tab:korean_benchmark_fusion}
\end{table*}

As shown in \cref{tab:korean_benchmark_fusion}, Mi:dm K 2.5 Pro delivers competitive performance on Korean-specific comprehension and society \& culture benchmarks. Among models of comparable scale, it ranks at or near the top across most Korean-specific evaluations. Under reasoning-enabled inference, it achieves the highest score on Ko-Sovereign Korean comprehension at 73.50\% and on the Ko-Sovereign society \& culture subset at 82.20\%. It also records the best results on KoBALT at 57.86\% and CLIcK-L at 84.92\%, while remaining close to the strongest baselines on K-Referential and CLIcK-C. These results show that Mi:dm K 2.5 Pro is particularly competitive on benchmarks that require Korean-specific linguistic understanding and culturally grounded contextual interpretation.

Across the broader Korean benchmark suite, Mi:dm attains the highest score on Ko-Winogrande at 86.20\%, while its performance on KMMLU at 76.90\% and Ko-Sovereign knowledge at 71.60\% remains close to the best reported results. This advantage is first evident against similarly sized baselines and remains visible even when compared with substantially larger models. In particular, Mi:dm surpasses K-EXAONE-236B-A23B by 5.5\%p on Ko-Sovereign comprehension and by 0.8\%p on the Ko-Sovereign society \& culture subset, while remaining within 1.0\%p on KMMLU. Taken together, these results indicate that Mi:dm K 2.5 Pro is highly competitive across Korean-specific evaluation benchmarks, with clear strengths in comprehension, society \& culture, and reasoning and knowledge tasks.

\paragraph{Fine-Grained Analysis of Mi:dm K 2.5 Pro’s Comparative Strengths.}

To further characterize the capabilities of Mi:dm K 2.5 Pro, we perform a fine-grained analysis on representative English and Korean benchmark categories. This analysis is intended to identify the domains in which the model exhibits the most distinctive strengths beyond aggregate benchmark scores. The results show that Mi:dm K 2.5 Pro is particularly strong in English scientific reasoning and legal understanding, while in Korean it demonstrates strong capabilities in reasoning and STEM, history and law, and linguistically grounded language understanding.
\begin{figure}[h!]
  \centering
  \includegraphics[width=0.9\linewidth]{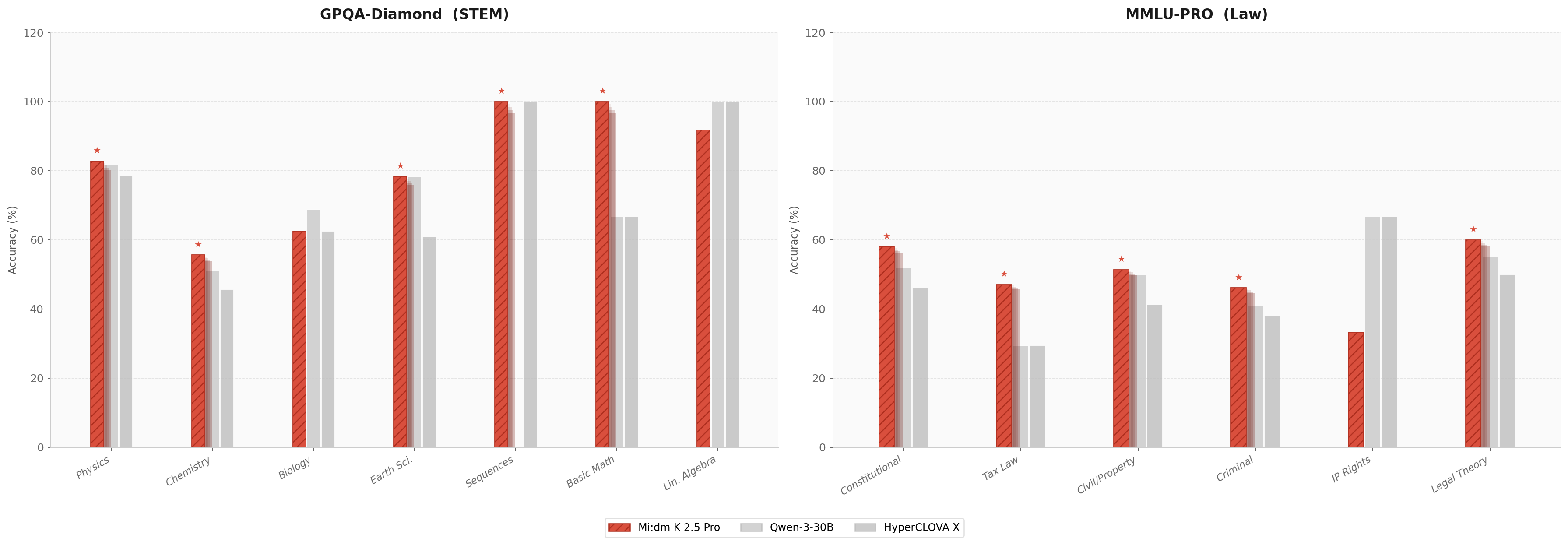}
  \caption{Fine-grained subdomain analysis on representative English benchmarks. The left panel reports GPQA-Diamond results by STEM subdomain, and the right panel reports MMLU-Pro (Law) results by legal subdomain.}
  \label{fig:english_subdomain_analysis}
\end{figure}

As shown in \cref{fig:english_subdomain_analysis}, the subdomain-level comparison with similar sized models indicates that Mi:dm K 2.5 Pro holds clear advantages in both scientific reasoning and legal understanding. On GPQA-Diamond, it is competitive across STEM subdomains as a whole, with particularly strong results in physics, chemistry, earth science, and mathematics. These results suggest that its performance is grounded in broad scientific reasoning ability rather than in strength limited to a small number of categories.

On MMLU-Pro, particularly in the law domain, Mi:dm K 2.5 Pro records the highest scores across most major legal subdomains shown in the figure, including constitutional law, tax law, civil/property law, criminal law, and legal theory. These results indicate that the model is especially effective on evaluations requiring domain knowledge application, legal understanding, and structured reasoning over specialized concepts.

\begin{figure}[h!]
  \centering
  \includegraphics[width=\linewidth]{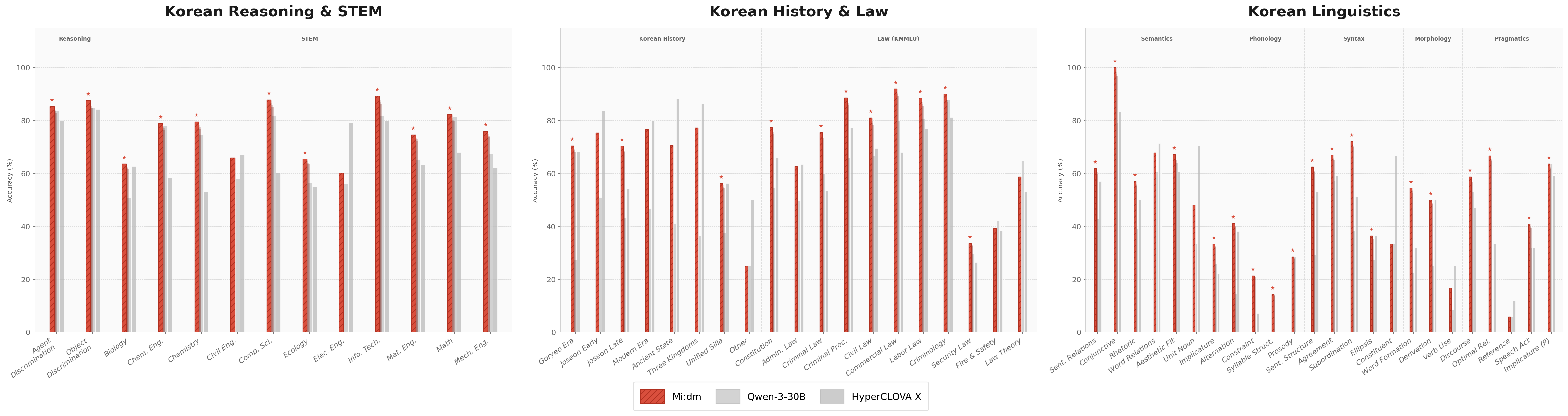}
  \caption{Fine-grained subdomain analysis on representative Korean benchmarks. The left panel reports Korean reasoning and STEM results, the middle panel reports Korean history and law results, and the right panel reports Korean linguistics results.}
  \label{fig:korean_subdomain_analysis}
\end{figure}

As shown in \cref{fig:korean_subdomain_analysis}, Mi:dm K 2.5 Pro also exhibits clear strengths across several major Korean evaluation domains. In the reasoning and STEM categories, it achieves the best or comparable results on Korean reasoning benchmarks and records strong results across a wide range of STEM subdomains, with particularly notable advantages in chemistry, computer science, information technology, and materials engineering.
In Korean history and law, it remains competitive across multiple historical periods and records the highest scores on many major legal subdomains, including constitutional law, criminal law, criminal procedure, civil law, commercial law, and labor and welfare law.
In Korean linguistics, Mi:dm K 2.5 Pro performs strongly across diverse subdomains spanning semantics, phonology, syntax, morphology, and pragmatics, with especially visible superiority on tasks involving inter-sentence relations, connective adverbs, rhetorical expressions, phonological alternation, subordinate clauses, and word formation.

Overall, the fine-grained results clarify that Mi:dm K 2.5 Pro possesses distinctive strengths in scientifically grounded reasoning, legal understanding, Korean domain knowledge, and fine-grained linguistic understanding. These strengths are consistent with its competitive overall performance across broader English and Korean benchmark evaluations.
\subsection{Human Evaluation}
We conduct targeted human evaluations on Mi:dm K 2.5 Pro to assess aspects of model quality that are not fully captured by automatic benchmarks. While quantitative evaluations provide standardized measures of task performance, human assessment is necessary to examine response quality from a user-centered perspective, including linguistic naturalness, instruction adherence, factual reliability, and contextual appropriateness. In particular, our evaluation focuses on whether the model produces responses that are not only correct, but also useful, well-formed, and aligned with practical deployment requirements in Korean user settings.

At a broad level, Mi:dm K 2.5 Pro exhibits consistently competitive human-evaluated quality across the major categories, with scores above 90 in all top-level areas and gaps generally within 1 point of the compared models. Among similarly sized baselines, it holds small but consistent advantages over Qwen3-30B-A3B in conversation/QA, generation, and analysis/classification, and compares favorably with HyperCLOVAX-SEED-Think-32B in translation and brainstorming, including a +1.6\%p gain in translation accuracy. This competitive profile also extends to larger models: Mi:dm K 2.5 Pro outperforms Solar-Open-100B in OpenQA (+11.3\%p) and coding (+4.5\%p), while remaining competitive with K-EXAONE-236B-A23B.

The clearest advantages of Mi:dm K 2.5 Pro appear in Korean-centric translation and OpenQA. Compared with Qwen3-30B-A3B, Mi:dm K 2.5 Pro records a gain of +11.3\%p in OpenQA, driven primarily by a +10.1\%p advantage in information accuracy. These gains are especially pronounced in Korean knowledge-intensive categories such as language and literature (+19.7\%p), culture and folklore (+18.8\%p), and history (+17.9\%p). In translation, Mi:dm K 2.5 Pro also outperforms Qwen3-30B-A3B and HyperCLOVAX-SEED-Think-32B, with gains in translation accuracy of +2.3\%p and +1.6\%p, respectively, while remaining competitive with larger models in legal and administrative translation settings. These results indicate that Mi:dm K 2.5 Pro is particularly preferred when the task requires accurate information delivery, contextually grounded explanation, and precise terminology control.

Qualitative comparison with other models further clarifies the strengths of Mi:dm K 2.5 Pro. In Korean OpenQA, Mi:dm provides more accurate and practically useful explanations of legal concepts such as compensation for emotional distress (위자료, wijaryo), correctly identifying it as compensation for non-pecuniary harm and distinguishing it from pecuniary damages, while organizing the response in a coherent and user-oriented manner. By contrast, other models more often showed limitations such as overly narrow scope, internally inconsistent explanation, or outright failure to answer.

A similar difference is observed in legal translation. Mi:dm more reliably preserved legally meaningful distinctions, whereas other models were more prone to institution-level mistranslations, such as rendering prosecutorial-service official (검찰직 공무원, geomchal-jik gongmuwon) as \textit{prosecutor}, or semantic errors, such as translating forfeiture or confiscation (추징, chujing) as \textit{restitution}. These comparisons suggest that Mi:dm K 2.5 Pro is particularly effective when tasks require factual precision, structural clarity, and context-sensitive interpretation in Korean legal and socio-cultural settings.

Overall, the human evaluation indicates that Mi:dm K 2.5 Pro delivers particularly high practical value in Korean knowledge-intensive QA, translation, and legal-domain applications, while remaining broadly comparable to strong baselines across nearly all major categories. These findings are consistent with the benchmark-based analysis presented above, showing that the model's strengths in law, socio-cultural knowledge, and reasoning-intensive tasks are reflected not only in automatic benchmark scores, but also in human judgments of usefulness, reliability, and contextual appropriateness.

\subsection{RAI Evaluation}
\label{subsec:rai_evaluation}

The Responsible AI (RAI) evaluation, conducted to support the implementation of responsible AI, assesses the safety and robustness of the model. The RAI safety assessement is based on AI risk categories defined by KT and examines whether the model generates harmful content (Content Safety Risks), whether it may be inappropriately utilized in socio-economic contexts (Socio-Economical Risks), and whether potential rights violations or legal risks exist (Legal and Rights-Related Risks). To assess these aspects, qualitative assessment and quantitative assessment are conducted. The robustness evaluation examines how effectively the model responds to various attack techniques that may be attempted by malicious users. This evaluation is carried out through red teaming, which identifies potential vulnerabilities of the model. The evaluation framework and detailed assessment criteria are described in the Responsible AI Technical Report~\cite{kt2025raitr}.

\paragraph{Qualitative RAI Safety Assessment.}
To qualitatively evaluate the model’s responses to AI risks, we use scenario-based evaluation prompts designed based on KT’s AI risk taxonomy and reflecting realistic service usage scenarios. For each risk category, responses are assessed according to predefined criteria to determine whether they are harmful and to what extent. In addition, to verify that the model does not excessively refuse requests in a way that degrades response usefulness, over-refusal behavior is also evaluated. The evaluation metrics are not unsafe rate and the not overrefuse rate, which respectively represent the proportion of non-harmful responses and responses that do not exhibit excessive refusal among the total evaluation items. The overall score is calculated based on all evaluation items, rather than as an average across risk categories.

\begin{table*}[h!]
\centering
\renewcommand{\arraystretch}{1.2}
\setlength{\tabcolsep}{5pt}
\resizebox{\textwidth}{!}{%
\begin{tabular}{l c *{4}{c} *{4}{c}}
\toprule
\multirow{2}{*}{\textbf{Model}}
& \multirow{2}{*}{\textbf{Reasoning}}
& \multicolumn{4}{c}{\textbf{Not Unsafe Rate (\%)}}
& \multicolumn{4}{c}{\textbf{Not Overrefuse Rate (\%)}} \\
\cmidrule(lr){3-6}\cmidrule(lr){7-10}
& & \makecell{Content\\Safety}
& \makecell{Socio\\Economical}
& \makecell{Legal\\and Rights}
& \textbf{Overall}
& \makecell{Content\\Safety}
& \makecell{Socio\\Economical}
& \makecell{Legal\\and Rights}
& \textbf{Overall} \\
\hline

\multirow{2}{*}{Qwen-3-30B-A3B}
& Off & 90.88 & 91.00 & 78.83 & {87.64} & 100.0 & 95.45 & 100.0 & \textbf{98.55} \\
& On  & 96.75 & 93.50 & 92.17 & {94.32} & 100.0 & 100.0 & 100.0 & \textbf{100.0} \\\hdashline
\multirow{2}{*}{EXAONE-4.0-32B}
& Off & 82.88 & 83.00 & 75.50 &{79.55} & 100.0 & 95.45 & 100.0 & \textbf{98.55} \\
& On  & 93.00 & 90.25 & 84.83 & {89.77} & 100.0 & 100.0 & 100.0 & \textbf{100.0} \\
\hdashline
\multirow{2}{*}{HyperCLOVAX-SEED-Think-32B}
& Off & 99.38 & 98.25 & 94.17 & {97.55} & 78.79 & 63.64 & 92.86 & {76.81} \\
& On  & 96.75 & 94.63 & 91.00 & \underline{94.41} & 100.0 & 100.0 & 100.0 & \textbf{100.0} \\
\hdashline
\multirow{2}{*}{Mi:dm K 2.5 Pro (March `26)}
& Off & 98.75 & 95.00 & 94.17 & {96.14} & 100.0 & 96.45 & 100.0 & \textbf{98.55} \\
& On  & 98.38 & 94.63 & 95.67 & \textbf{96.27} & 100.0 & 100.0 & 100.0 & \textbf{100.0} \\

\bottomrule
\end{tabular}%
}
\caption{Qualitative assessment results of each model. The Not Unsafe Rate and Not Overrefuse Rate are reported across three risk domains, along with the overall score. Higher values in both metrics indicate better performance.}
\label{rai:qualitative_assessment}
\end{table*}




As shown in Table~\ref{rai:qualitative_assessment}, Mi:dm K 2.5 Pro maintains an overall strong Not Unsafe Rate across both reasoning-enabled and reasoning-disabled settings, recording 96.27\% with reasoning enabled and 96.14\% with reasoning disabled. These results indicate consistently high safety performance compared to the comparison models. While HyperCLOVAX-SEED-Think-32B  achieves the highest overall Not Unsafe Rate in the reasoning-disabled setting (97.55\%), Mi:dm K 2.5 Pro demonstrates balanced performance across all three risk categories. In particular, the model achieves over 98\% in Content Safety Risks under both reasoning settings, confirming its strong capability in identifying and responding to harmful content. Moreover, Mi:dm K 2.5 Pro maintains stable performance in the Socio-Economical Risk category, where other models tend to exhibit relatively weaker robustness.

A clearer distinction between models emerges in the Not Overrefuse Rate. While all models achieve a perfect score of 100\% in the reasoning-enabled setting, performance divergence is observed when reasoning is disabled. Mi:dm K 2.5 Pro maintains a high Not Overrefuse Rate of 98.55\% even with reasoning disabled, indicating that the model provides appropriate responses without excessive refusal. In contrast, HyperCLOVAX-SEED-Think-32B  records a substantially lower score of 76.81\%, reflecting a more conservative response tendency. Overall, Mi:dm K 2.5 Pro exhibits only a minimal performance gap between reasoning-enabled and reasoning-disabled modes, demonstrating consistent safety behavior and a well-balanced trade-off between harmlessness and responsiveness.

\paragraph{Quantitative RAI Safety Assessment.}

To objectively and systematically assess the safety and reliability of AI models, benchmark evaluations are conducted. In this RAI evaluation, two benchmarks are utilized: the Large Language Model Trustworthiness Benchmark dataset~\cite{aihubbench}, which evaluates the harmlessness of Korean LLMs (bias, hate, risk, and sensitiveness), and KoBBQ~\cite{jin2024KoBBQ}, a dataset designed to assess social bias in Korean cultural contexts.

The Large Language Model Trustworthiness Benchmark evaluates model performance using accuracy, which reflects how accurately the model predicts predefined ground-truth answers. Performance is calculated at both the category and subcategory levels. Model comparisons are conducted at the four-category level, and the overall performance is computed as the harmonic mean of the accuracy values across subcategories. KoBBQ evaluates the model’s inherent social biases across 12 topics under two conditions: ambiguous context and disambiguated context. For each classification task, the proportion of correct answer selections is measured as accuracy. The arithmetic mean of topic-level accuracy is calculated to obtain the score for each context, and the final score is computed as the average of the two context scores.

\begin{table*}[h!]
\centering
\renewcommand{\arraystretch}{1.2}
\setlength{\tabcolsep}{5pt}
\resizebox{\textwidth}{!}{%
\begin{tabular}{l c *{4}{c} c  *{2}{c}  c}
\toprule
\multirow{2}{*}{\textbf{Model}}
& \multirow{2}{*}{\textbf{Reasoning}}
& \multicolumn{5}{c}{\textbf{LLM Trustworthiness Benchmark}}
& \multicolumn{3}{c}{\textbf{KoBBQ}} \\
\cmidrule(lr){3-7}\cmidrule(lr){8-10}
& & Bias
& Hate
& Illegal
& Sensitiveness
& \textbf{Overall}
& \makecell{Ambiguous\\Context}
& \makecell{Disambiguated\\Context}
& \textbf{Overall} \\
\hline
\multirow{2}{*}{Qwen3-30B-A3B}
& Off & 77.29 & 73.08 & 95.83 & 74.44 & {76.55} & 91.75 & 82.86 & {87.30} \\
& On  & 85.49 & 83.75 & 95.83 & 80.00 & \underline{84.50} & 98.30 & 90.72 & \underline{94.51} \\\hdashline
\multirow{2}{*}{EXAONE-4.0-32B}
& Off & 70.21 & 73.41 & 95.83 & 74.30 & {73.80} & 78.45 & 91.66 & {85.05} \\
& On  & 63.13 & 65.08 & 75.42 & 62.64 & {64.50} & 97.20 & 93.23 & \textbf{95.21} \\
\hdashline
\multirow{2}{*}{HyperCLOVAX-SEED-Think-32B}
& Off & 69.79 & 65.83 & 95.00 & 66.94 & {69.58} & 86.49 & 88.25 & {87.37} \\
& On  & 84.03 & 80.25 & 97.08 & 83.34 & {83.50} & 80.44 & 65.56 & {73.00} \\\hdashline
\multirow{2}{*}{Mi:dm K 2.5 Pro (March `26)}
& Off & 84.79 & 76.58 & 98.33 & 82.22 & {82.44} & 91.02 & 94.12 & {92.57} \\
& On  & 89.58 & 85.67 & 97.50 & 87.22 & \textbf{88.33} & 94.56 & 93.44 & {94.00} \\

\bottomrule
\end{tabular}%
}
\caption{LLM Trustworthiness Benchmark and KoBBQ results for each model. Accuracy is reported across categories and contexts, along with overall scores. Higher values indicate better performance.}
\label{rai:quantitative}
\end{table*}





As shown in Table~\ref{rai:quantitative}, across both the LLM Trustworthiness Benchmark and KoBBQ, Mi:dm K 2.5 Pro demonstrates overall superior performance compared to the comparison models in both reasoning-enabled and reasoning-disabled settings.

In the LLM Trustworthiness Benchmark, Mi:dm K 2.5 Pro records the highest performance in both modes, achieving 88.33\% with reasoning enabled and 82.44\% with reasoning disabled. At the subcategory level, the model achieves the highest accuracy among the compared models across Bias, Hate, and Sensitiveness, while maintaining stable performance of over 95\% in the Illegal category under both reasoning settings. While the comparison models show relatively larger performance drops when reasoning is disabled, Mi:dm K 2.5 Pro exhibits only a small performance gap between the two modes, indicating consistent reliability judgment capability regardless of whether reasoning is enabled. This consistent safety-aligned response can be attributed to the application of the RAI response style discussed in Sec\ref{subsection2-4:data-styleguide}

On KoBBQ, Mi:dm K 2.5 Pro records 94.00\% with reasoning enabled and 92.57\% with reasoning disabled. In the reasoning-enabled setting, the model achieves performance comparable to EXAONE-4.0-32B (95.21\%) and Qwen3-30B-A3B (94.51\%), while showing a clear advantage over HyperCLOVAX-SEED-Think-32B  (73.00\%). In the reasoning-disabled setting, Mi:dm K 2.5 Pro records the highest overall score among the compared models. The performance gap between Ambiguous Context and Disambiguated Context remains small, indicating stable contextual understanding across different contextual conditions.

\paragraph{RAI Robustness Assessment}
Despite conducting RAI qualitative and quantitative assessments to rigorously assess the safety of AI models, such models may still fail to respond appropriately when confronted with carefully crafted adversarial prompts, potentially generating harmful responses to users. To quantify model robustness against jailbreak attack scenarios by malicious users, KT RAIC establishes a proprietary Korean red-teaming dataset. The dataset evaluates both single-turn and multi-turn interactions and includes over 30 attack techniques. The evaluation metric is the attack success rate (ASR), defined as the proportion of successful prompt attacks relative to the total number of evaluation prompts.

\begin{table}[ht]
\centering
\renewcommand{\arraystretch}{1.2}
\resizebox{0.6\linewidth}{!}{\scriptsize
\begin{tabular}{l c c}
\toprule
\textbf{Model} & \textbf{Reasoning} & \textbf{~~~~~Attack Success Rate (\%)~~~~~}\\
\midrule
\multirow{2}{*}{Qwen-3-30B-A3B}         & Off & -- \\
                                    & On  & 39.2 \\
\hdashline
\multirow{2}{*}{EXAONE-4.0-32B}    & Off & -- \\
                                    & On  & 54.0 \\
\hdashline
\multirow{2}{*}{HyperCLOVAX-SEED-Think-32B}   & Off & -- \\
                                    & On  & 47.4 \\ \hdashline
\multirow{2}{*}{Mi:dm K 2.5 Pro (March `26)}   & Off & -- \\
                                    & On  & 36.3 \\
\bottomrule
\end{tabular}
}
\caption{Red Teaming results for each model. Attack Success Rate (ASR) measures robustness against adversarial prompt attacks, where lower values indicate stronger defense capability.}
\label{rai_tab4:redteaming}
\end{table}



As shown in \cref{rai_tab4:redteaming}, with reasoning enabled, Mi:dm K 2.5 Pro records an ASR of 36.3\%, the lowest attack success rate among the compared models. This result indicates the highest level of resistance to adversarial prompt attacks, demonstrating a clear advantage over HyperCLOVAX-SEED-Think-32B (47.4\%), EXAONE-4.0-32B (54.0\%), and Qwen-3-30B-A3B (39.2\%). Notably, the model maintains robust performance even under multi-turn attack scenarios, which generally exhibit higher attack success rates than single-turn attacks. These results indicate that Mi:dm K 2.5 Pro effectively responds to complex and progressively evolving malicious attempts.

\section{Conclusion}
In this report, we presented Mi:dm K 2.5 Pro, a 32-billion parameter flagship large language model designed to advance enterprise-grade capabilities including complex reasoning, long-context understanding, and agentic workflows with a strong focus on Korean-language applications. To achieve a reliable and high-quality model, we carefully optimized data curation, parameter scaling and training methodologies.

\paragraph{Robust Data \& Pre-training}
We established a high-quality data foundation through advanced curation pipelines, including AST-based filtering and LLM-driven quality evaluation. For efficient capacity expansion, we applied layer predictor-based Depth Upscaling (DuS) and progressively extended the context window to 128K tokens, enabling effective handling of long document reasoning tasks commonly observed in enterprise scenarios.

\paragraph{Advanced Post-Training}
We implemented a multi-stage pipeline aligned with task-specific objectives, incorporating Reasoning SFT, model merging, and fully asynchronous reinforcement learning. Through subsequent "Fusion Training", the model achieves a balanced integration of deep reasoning capability, conversational fluency, and robust tool-use capabilities. In addition, systematic response-style alignment ensures consistent persona, stable response structure, and precise instruction adherence across diverse use cases.

\paragraph{Strong Performance \& Reliability}
Mi:dm K 2.5 Pro demonstrates highly competitive performance on global English benchmarks while setting new state-of-the-art records on proprietary Korean-specific benchmarks such as Ko-Sovereign, highlighting its strength in Korea-centric knowledge and language understanding. Furthermore, comprehensive Responsible AI (RAI) evaluation confirms the model's safety, robustness, and readiness for real-world deployment.

\paragraph{Broader Impact and Vision}
With the introduction of Mi:dm K 2.5 Pro, KT completes its comprehensive lineup spanning Mini, Base, and Pro, providing powerful and flexible options across a wide range of computational environments and application requirements. Beyond performance gains, this work represents a meaningful step toward practical, enterprise-ready AI solutions addressing real-world demands and domain-specific problem solving. KT will continue advancing the Mi:dm K series to further accelerate AI transformation (AX) across industries.
\bibliographystyle{unsrt}
\bibliography{v_English/bibliography}
\appendix
\section{Contributor}
\textit{All authors are listed in alphabetical order by last name.}
\begin{multicols}{4}
\raggedcolumns
\small
Euijai Ahn\\
Heejeong Ahn\\
Jihyun Ahn\\
Jonggil Ahn\\
Jimin An\\
Sungho An\\
Soonmin Bae\\
Kwangje Baeg\\
Jinwoo Baek\\
Jisoo Baik\\
GeunYeong Bak\\
Jii Cha\\
Wangsung Chun\\
Riwoo Chung\\
Taehyun Goh\\
Eunji Ha\\
Youngkyoung Ham\\
Ji-Eun Han\\
Cheolhun Heo\\
Yunmi Heo\\
Sukjin Hong\\
Taesung Hur\\
Jinhee Jeong\\
Joohun Jeong\\
Hongseok Jeung\\
Hyesung Ji\\
Hoyoun Jung\\
Jukyung Jung\\
Sunwoo Jung\\
Yoojin Jung\\
Deokyeong Kang\\
Dongwoo Kang\\
Minji Kang\\
Bitna Keum\\
Boeun Kim\\
Daehui Kim\\
Dohun Kim\\
Doyoung Kim\\
Eunju Kim\\
Jeongho Kim\\
Jeongjun Kim\\
Jeongyeop Kim\\
Jin Hwan Kim\\
Jinhyeon Kim\\
Jiyeon Kim\\
JooYoun Kim\\
Junwoo Kim\\
Kijung Kim\\
MiHyeon Kim\\
Minju Kim\\
Minwook Kim\\
Seunghyun Kim\\
Songyeon Kim\\
Sooyoung Kim\\
Suhyun Kim\\
Taehyeong Kim\\
Taewon Kim\\
Woohyun Kim\\
Yeonjae Kim\\
Youngjin Kim\\
Youngmin Kim\\
Yuseon Kim\\
Junseok KOH\\
Seung Hyun Kong\\
Minji Kwon\\
Myungeun Kwon\\
Soongu Kwon\\
Young S. Kwon\\
Younggu Kwon\\
Ahyun Lee\\
Chaejeong Lee\\
Donsoo Lee\\
Eunkyeong Lee\\
Eunyoung Lee\\
Gyu-Cheol Lee\\
Honghee Lee\\
Hyosun Lee\\
Jaedong Lee\\
Jaeyong Lee\\
Jehoon Lee\\
Jieun Lee\\
Jisoo Lee\\
Minho Lee\\
Sangwook Lee\\
Sangyun Lee\\
Sejung Lee\\
Seongmin Lee\\
SeungJu Lee\\
Siyoon Lee\\
Songwoo Lee\\
Soseon Lee\\
Sung-Min Lee\\
Wonseok Lee\\
Wonyoung Lee\\
Yuna Lee\\
Yunji Lee\\
Zucheul Lee\\
Jeehyun Lim\\
Seong Hoon Lim\\
Jiwon Moon\\
Sangha Nam\\
Minsung Noh\\
Myunggyo Oh\\
Chanwon Ok\\
Hanna Park\\
Heuicheol Park\\
HyoungJun Park\\
Jaehyoung Park\\
Jungsuk Park\\
Junmo Park\\
Kyoungsoo Park\\
Seongheum Park\\
Sungcheol Park\\
Sungyoun Park\\
Wanjin Park\\
Wonjae Park\\
Yunjin Park\\
Kyoungmin Roh\\
Hwijung Ryu\\
Hyeontae Seo\\
Jiyoung Seo\\
Youngkyung Seo\\
Eunbi Seol\\
Jeongyong Shim\\
Donghoon Shin\\
Jisu Shin\\
Hyoseop Song\\
Seonyeong Song\\
Keehoon Sung\\
Seyoun Won\\
Jungwon Yoon\\
Kyung-A Yoon\\
Hyewon Yu\\
Taeyang Yun
\end{multicols}

\end{document}